%% file: main.tex
\documentclass[twoside]{article}
\usepackage[accepted]{format}

\usepackage[hidelinks]{hyperref}
\usepackage{natbib}
\usepackage{caption}
\usepackage{booktabs, multirow}
\usepackage{url}

\usepackage{amsmath}
\usepackage{amssymb}
\usepackage{mathtools}
\usepackage{amsthm}
\usepackage{xspace}
\usepackage{bm}
\usepackage{bbm}
\usepackage{enumitem}
\usepackage{diagbox}
\usepackage{xr}

\usepackage[left=1.5cm, right=1.5cm, top=2cm, bottom=2cm]{geometry}

\theoremstyle{plain}
\newtheorem{theorem}{Theorem}[section]

\newtheorem{lemma}[theorem]{Lemma}

\theoremstyle{definition}

\theoremstyle{remark}

\newcommand{\eps}{\varepsilon}

\newcommand{\E}{\mathbb{E}}

\newcommand{\R}{\mathbb{R}}
\newcommand{\Z}{\mathbb{Z}}
\newcommand{\textbfne}[1]{\textbf{\fontseries{b}\selectfont #1}}

\newcommand{\mc}[1]{\ensuremath{\mathcal{#1}}\xspace}
\newcommand{\mb}[1]{\ensuremath{\mathbf{#1}}\xspace}

\newcommand{\tn}[1]{\ensuremath{\textnormal{#1}}\xspace}
\newcommand{\ol}[1]{\ensuremath{\overline{#1}}\xspace}

\DeclareMathAlphabet\mathbfcal{OMS}{cmsy}{b}{n}

\newcommand{\mbc}[1]{\ensuremath{\mathbfcal{#1}}\xspace}

\newcommand{\bx}{{\mathbf{x}}}

\newcommand{\bz}{{\mathbf{z}}}
\newcommand{\bu}{{\mathbf{u}}}

\newcommand{\br}{{\mathbf{r}}}

\begin{document}

\twocolumn[

\papertitle{Learning from Label Proportions and Covariate-shifted Instances} 
\paperauthor{ Sagalpreet Singh \And Navodita Sharma$^*$ \And  Shreyas Havaldar$^*$ }
\authoraddress{ Google DeepMind \And  Google DeepMind \And Google DeepMind }
\paperauthor{ Rishi Saket \And Aravindan Raghuveer }
\authoraddress{ Google DeepMind \And Google DeepMind }]

\begin{abstract}
  In many applications, especially due to lack of supervision or privacy concerns, the training data is grouped into bags of instances (feature-vectors) and for each bag we have only an aggregate label derived from the instance-labels in the bag. In learning from label proportions (LLP) the aggregate label is the average of the instance-labels in a bag, and a significant body of work has focused on training models in the LLP setting to predict instance-labels.
  In practice however, the training data may have fully supervised albeit covariate-shifted \emph{source} data, along with the usual \emph{target} data with bag-labels, and we wish to train a good instance-level predictor on the target domain.
  We call this the covariate-shifted hybrid LLP problem.
  Fully supervised covariate shifted data often has useful training signals and the goal is to leverage them for better predictive performance in the hybrid LLP setting.
To achieve this, we develop methods for hybrid LLP which naturally incorporate the target bag-labels along with the source instance-labels, in the domain adaptation framework. Apart from proving theoretical guarantees bounding the target generalization error, we also conduct experiments on several publicly available datasets showing that our methods outperform LLP and domain adaptation baselines as well techniques from previous related work.

\end{abstract}
\input{introduction}

\input{previous}

\input{preliminaries}

\input{proposed}

\input{experiments}

\input{results}

\input{conclusion}

\bibliographystyle{apalike}
\bibliography{references}

\appendix

\input{supplement}
\end{document}

%% file: introduction.tex
\section{Introduction}\label{sec:intro}
Learning from label proportions (LLP) is a direct generalization of supervised learning where the training instances (i.e., feature-vectors) are partitioned into \textit{bags} and for each bag only the average label of its instances is available as the \emph{bag}-label. Full supervision is equivalent to the special case of unit-sized bags. In LLP, using  bags of instances and their bag-labels, the goal is to train a good predictor of the instance-labels. Over the last two decades, LLP has been used in scenarios with lack of fully supervised data due to legal requirements~\citep{R10}, privacy constraints~\citep{WIBB}  or coarse supervision~\citep{CHR}. Applications of LLP include image classification~\citep{Bortsova18, Orting16}, spam detection~\citep{QSCL09}, IVF prediction~\citep{hernandez2018}, and high energy physics~\citep{DNRS}. 
More recently, restrictions on cross-site tracking of users has led to coarsening of previously available fine-grained signals which have been used to train large-scale models predicting user behavior for e.g. clicks or product preferences. Popular  mechanisms  (see Apple SKAN~\citep{skan} and Chrome Privacy sandbox~\citep{sandbox}) aggregate relevant labels for bags of users resulting in LLP training data. Due to revenue criticality of user modeling in advertising, the study of LLP specifically for such applications has gained importance. A popular baseline method to train models using training bags and their bag-labels is to minimize a bag-level loss which for any bag is some suitable loss function between the the average prediction and the bag-label (see \cite{ArdehalyC17}). Other methods using different bag-level losses have also been proposed (e.g. \cite{OT,fast-llp}) for training models in the LLP setting.

One aspect of data in real-world applications is its heterogeneity, which introduces new aspects to the vanilla LLP modeling formulation. In particular, apart from bag-level data from the \textit{target} distribution, the learner may have access to instance labels from a covariate-shifted \emph{source} distribution. For example, in  user behavior modeling for online advertising, while bag-level aggregate labels could be available for a target set of (privacy sensitive) users as mentioned above, other users may choose to share browsing and purchase history, which would yield covariate-shifted source data with instance-level labels. This is also mentioned in Section 2.1 of \cite{obrien2022challengesapproachesprivacypreserving} which states: ``\emph{.. some platforms may continue to allow conversion tracking, and some users may also choose to allow conversion tracking , the training set is likely to contain some examples with individual labels and some examples with only group labels''}.
This can also occur when the source originates from geographies which impose less stringent privacy constraints on data corresponding to online activity, medical records or financial transactions, thereby not requiring the aggregation of labels. Recent work has also studied age-dependent privacy, in which \emph{releasing outdated data may lead to less privacy leakage if a user only focuses on protecting its real-time status} (from Section 1 of \cite{AgeDependentDiff}, see also \cite{AgeAwareDiff}). Such outdated data could correspond to the source distribution for which instance-labels are available. 

Here, we think of covariate-shift as a difference in $p(\mb{X})$ i.e. the distribution of feature-vectors, between the source $\mc{D}_S$ and target $\mc{D}_T$ distributions, with the conditional label distribution $p(Y\mid\mb{X})$ being the same on $\mc{D}_S$ and $\mc{D}_T$. We call this \emph{covariate-shifted hybrid LLP} in which the goal is to leverage the full supervision on the source as well as the bag-level supervision on the target to train better instance-label predictors on the target distribution. 

 Previous works \citep{Domain-Adaption-AC,Li-Culotta} studied the case where the source training data was aggregated into bags whose bag-labels are available, while the training data from the target distribution is completely unsupervised. The work of \cite{Domain-Adaption-AC} gave a \emph{self-training} based approach where the model trained on the source data is used to predict bag-labels on the unsupervised target train-set from which a subset of the most confidently labeled bags are used (along with the source data) to retrain the predictor. The more recent work of \cite{Li-Culotta} proposed solutions directly applying domain adversarial neural-network (DANN) methods in which apart from minimizing the bag-level loss on the source data, an unsupervised domain prediction loss is \emph{maximized} to ensure that the predictor is domain-independent. %

The works of  \cite{Domain-Adaption-AC}, and \cite{Li-Culotta} as well as standard domain adaptation methods (e.g. \cite{long2015learning}) can be applied to our setting by simply ignoring the bag-labels of the target train-set, and treating the labeled instances in the source data as bags of size $1$.  Note however that these approaches discard the informative signal from the target bag-labels and are thus likely to degrade the predictive performance.

The main contributions of this paper are a suite of techniques which use the bag-labels from the target training set, not only to minimize the bag-loss i.e., the predictive loss on bags, but also to do better domain adaptation. We focus on regression as the underlying task and propose loss functions which, at a high level, have three components: (i) the instance-level loss on the source data, (ii) a bag-level loss on the target training bags, and (iii) a domain adaptation loss which leverages the instance-labels from source and bag-labels from target. Our main methodological novelty is the third term which leverages bag-labels (unlike previous works) from the target domain for domain adaptation, along with instance-labels from the source domain. 
Specifically, our BL-WFA method using the ${\sf BagCSI}$  loss (eqn. \eqref{eqn:BagCSI}) is the first to incorporate the instance-labels from the source along with the target bag-labels into the domain adaptation loss. The design of our ${\sf BagCSI}$ loss is theoretically justified: we prove generalization error bounds (Section \ref{sec:our_contrib}), %
and we also generalize this to PL-WFA which can use target-level pseudo-labels instead (see Section \ref{proposed} for details of BL-WFA and PL-WFA).
Complementing these analytical insights, we provide in Section \ref{sec:experiments} extensive experimental evaluations of our methods showing performance gains, on real as well as synthetic datasets.

%% file: previous.tex
\section{Previous Related Work}
{\it Learning from Label Proportions (LLP).} Early work on LLP by \cite{FreitasK05,HG} applied trained probabilistic models using Monte-Carlo methods, while \cite{Musicant,R10} provided adaptations of standard supervised learning approaches such as SVM, $k$-NN and neural nets, and \cite{ChenLQZ09,StolpeM11} developed clustering based methods for LLP. More specialized techniques were proposed by \cite{QuadriantoSCL09} and later extended by \cite{PatriniNCR14}  to estimate parameters from label proportions for the exponential generative model assuming well-behaved label distributions of bags. An optimization based approach of \cite{YuLKJC13} provided a novel $\propto$-SVM method for LLP. Newer methods involve deep learning~\citep{KDFS15,DZCBV19,LWQTS19,NSJCRR22} and others leverage characteristics of the distribution of bags~\citep{SRR,ZWS22,chen2023learning} while \cite{easy-llp} developed model training techniques on derived \emph{surrogate} labels for instances for random bags. Defining the LLP label proportion regression task in the PAC framework, \cite{YCKJC14}, established bounds on the generalization error bounds for bag-distributions. For the classification setting and specific types of loss functions, bag-to-instance generalization error bounds were shown by \cite{easy-llp, chen2023learning}.

{\it Domain Adaptation.} %
Many of the domain adaptation techniques try to align the source and target distributions by minimizing a distance-measure between domains. 
The work of \cite{long2015learning} generalized deep convolutional neural networks to the domain adaptation scenario, by matching the task-specific hidden representations for the source and target domains in a reproducing kernel Hilbert space. An extension of this work by \cite{long2017deep} proposed a Joint Adaptation Network which aligns the joint distributions of multiple domain-specific hidden layers using a joint maximum mean discrepancy measure. The technique proposed by \cite{ganin2016domain} focuses on learning from features which are indiscriminate with respect to the shift between domains. Recently, \cite{li2023domain} applied domain adaptation to the LLP and proposed a model combining domain-adversarial neural network (DANN) and label regularization, to learn from source-domain bags and predict on instances from a target domain.

%% file: preliminaries.tex
\section{Preliminaries}\label{sec:prelim}
For a given $d \in \Z^+$, feature-vectors (instances) are $d$-dimensional reals and labels are real-valued scalars.
Let $\mc{D}_S$ and $\mc{D}_T$ denote respectively the source and target distributions over $\R^d\times [0,1]$.

We denote by $\mc{S}(n)$ a source \emph{training} set of $n$ examples $\{(\bx_i, y_i)\,\mid\, i = 1,\dots, n\}$ drawn iid from $\mc{D}_S$, and analogously define $\mc{T}(n)$ as $n$ iid examples from $\mc{D}_T$. However, while the source training set is available at the instance-level, the target train-set is aggregated randomly into \emph{bags}. We specify the bag-creation as follows. \\
{\it Target Training Bags.} A bag $B\subseteq \R^d$ is a finite set of instances $\bx$ with labels $y_{\bx}$ and its \emph{bag-label} $y_B := (1/|B|)\sum_{\bx \in B}y_{\bx}$ is the average of the instance-labels in the bag. The sample target training bags denoted by $\mc{B}(m,k)$ is a random set of $m$ $k$-sized bags $(B_1, y_{B_1}), \dots, (B_m, y_{B_m})$ created as follows:
\begin{enumerate}[nolistsep,noitemsep]
    \item Let $\mc{T}(mk) :=  \{(\bx_i, y_i)\,\mid\, i = 1,\dots, mk\}$ be $mk$ iid examples from $\mc{D}_T$.
    \item Let $I_j = \{k(j-1) + 1, \dots, kj\}$, $j=1,\dots, m$ be a partition of $[mk]$. 
    \item For each $j = 1, \dots, m$, let $B_j = \{\bx_i\,\mid\, i \in I_j\}$ with bag-labels $y_{B_j} = (1/k)\sum_{i \in I_j}y_i$. 
\end{enumerate}
{\it Instance and Bag-level losses.} Since we focus on regression as the underlying task for an instance-level predictor we shall define our losses using \emph{mean squared-error} (mse). For any function $h : \R^d \to \R$, the loss w.r.t. to a distribution $\mc{D}$ over $\R^d\times \R$ is
\begin{equation*}
    \eps(\mc{D}, h) := \E_{(\bx,y)\leftarrow D}\left[\left(h(\bx) - y\right)^2\right],
\end{equation*}
where we shall let $\mc{D}$ be $\mc{D}_S$ or $\mc{D}_T$ for our purpose. The loss over a finite sample $\mc{U}$ of labeled points is:
\begin{equation*}
    \hat{\eps}(\mc{U}, h) := \frac{1}{|\mc{U}|}\sum_{(\bx,y)\in \mc{U}}\left[\left(h(\bx) - y\right)^2\right] 
\end{equation*}
where we shall take $\mc{U}$ as the source training-set $\mc{S}$ or target training-set $\mc{T}$ (we omit the sizes of the train-set for convenience). Finally, we have the loss on sampled bags:
\begin{eqnarray*}
    & & \bar{\eps}(\mc{B}, h) \nonumber \\ &:=& \frac{1}{|\mc{B}|}\sum_{(B, y_B) \in \mc{B}}\left[\left(\left(\frac{1}{|B|}\sum_{\bx \in B}h(\bx)\right) - y_B\right)^2\right]
\end{eqnarray*}

{\it Function Classes and pseudo-dimension.} We will consider a class $\mc{F}$ of real-valued functions (regressors) mapping $\R^d$ to  $[0, 1]$. For any $\mbc{X} \subseteq \R^d$ s.t. $|\mbc{X}| = N$, let $\mc{C}_p(\xi, \mc{F}, \mbc{X})$ denote a minimum cardinality $\ell_p$-metric $\xi$-cover of $\mc{F}$ over $\mbc{X}$, for some $\xi > 0$. Specifically, $\mc{C}_p(\xi, \mc{F}, \mbc{X})$ is a minimum sized subset of $\mc{F}$ such that for each $f^* \in \mc{F}$, there exists $f \in \mc{C}_p(\xi, \mc{F}, \mbc{X})$ s.t. $\left(\E_{\bx \in \mbc{X}}\left[\left|f^*(\bx) - f(\bx)\right|^p\right]\right)^{1/p} \leq \xi$ for $p \in [1,\infty)$, and $\max_{\bx\in \mbc{X}}\left|f^*(\bx) - f(\bx)\right| \leq \xi$ for $p =\infty$.

As detailed in Sections 10.2-10.4 of \cite{Anthony-Bartlett}, the largest size of such a cover over all choices of $\mbc{X} \subseteq \R^d$ s.t. $|\mbc{X}| = N$ is defined to be  $N_p(\xi, \mc{F}, N)$. %

The \emph{pseudo-dimension} of $\mc{F}$, ${\sf Pdim}(\mc{F})$ (see Section 10.4 and 12.3 of \cite{Anthony-Bartlett}, Appendix \ref{app:pseudo-dimension}) can be used to bound the size of covers for $\mc{F}$ as follows: %
\begin{equation}
    N_1(\xi, \mc{F}, N) \leq N_\infty(\xi, \mc{F}, N) \leq (eN/\xi p)^p \label{eqn:coversize}
\end{equation}
where $p = {\sf Pdim}(\mc{F})$ and $N \geq d$.

Since the task of our interest is regression, we shall assume that for any $f \in \mc{F}$, $f(\bx) = \br_f^{\sf T}\phi(\bx)$ where $\phi$ is a mapping to a real-vector in an embedding space and $\br_f$ is the representation of $f$ in that space (see Appendix \ref{app:simplifying} for an explanation).

\subsection{Our Contributions} \label{sec:our_contrib}
For $\mc{S} = \mc{S}(mk) = \{(\bz_i, \ell_i)\}_{i=1}^{mk}$, and $\mc{B} = \mc{B}(m, k) = \{(B_j, y_{B_j})\}_{j=1}^m$ be the bags constructed from $\mc{T} = \mc{T}(mk)$, we define the following \emph{covariate-shift} loss.
\begin{align}\label{eq:covariate_shifted_loss}
 \xi(\mc{S}, \mc{B}) & := 2\left\|\frac{1}{m}\sum_{j=1}^m y_{B_j}\left( \frac{1}{k}\sum_{\bx \in B_j}\phi(\bx)\right)\right. \nonumber \\ 
    & \qquad \qquad \qquad \qquad \left. - \frac{1}{mk}\sum_{i=1}^{mk}l_i\phi(\bz_i) \right\|_2
\end{align}
Note that the above domain adaptation loss depends on the labels from the source train-set labels as well as the bag-labels of the target training bags. In other words, it leverages the supervision provided on the training data $\mc{S}$ and $\mc{B}$. 
We bound the difference of the sample bag-loss on target training bags $\mc{B}$ and the sample instance-level loss on the source as follows.
\begin{lemma} \label{lem:main1} For any $h \in \mc{F}$,
\begin{equation*}
    \bar{\eps}(\mc{B}, h) - \hat{\eps}(\mc{S}, h) \leq \xi(\mc{S}, \mc{B})\left\|\br_h\right\|_2 + \lambda'(\mc{S}, \mc{T}) + R(h, \mc{S}, \mc{T})
\end{equation*}
where $\lambda'(\mc{S}, \mc{T})$ is independent of $h$ and $R(h, \mc{S}, \mc{T})$ is a label-independent regularization on $\mc{S}$ and $\mc{T}$.
\end{lemma}
The above  lemma whose proof along with the expressions for $\lambda'(\mc{S}, \mc{T})$ and  $R(h, \mc{S}, \mc{T})$, is provided in Section \ref{sec:lemmadiffbd}, shows that minimizing the instance-level loss on the source train-set $\mc{S}$ along with the covariate-shift loss training data can upper bound the bag-level loss on the target training bags $\mc{B}$. Since our goal is to upper bound the instance-level loss on the target distribution, we bound the latter using the bag-loss on the training bags in the following novel generalization error bound.

\begin{theorem}\label{thm:main1}
    For  $m, k \in \Z^+, \nu, \delta > 0$, w.p. $1-\delta$ over choice of $\mc{B} = \mc{B}(m,k)$,  $\eps(\mc{D}_T, h) \leq 16k\bar{\eps}(\mc{B}, h)$ for all $h \in \mc{F}$ s.t. $\eps(\mc{D}_T, h) \geq \nu$ and $p = {\sf Pdim}(\mc{F})$, when $m \geq
    O\left(\left(p\left(\log\left(\frac{k}{\nu}\right) + \log\log\left(\frac{1}{\delta}\right)\right) + \log\frac{1}{\delta}\right)\max\left\{\frac{1}{k\nu^2}, \frac{k^2}{\nu}\right\}\right)$.
\end{theorem}
The above is, to the best of our knowledge, the first bag-to-instance generalization error bound for regression tasks in LLP using the pseudo-dimension of the regressor class. Note however that there is a blowup in the error proportional to the bag-size $k$, which is understandable since, due to convexity, the mse loss between the average prediction in a bag and its bag-label is less than the average loss of the instance-wise predictions and labels. In other words, the error bound from Theorem \ref{thm:main1} is weaker with increasing bag size, and in Appendix \ref{app:error_bound_weakening} we demonstrate through an example that this  degradation with bag-size is unavoidable.

Lemma \ref{lem:main1} can, however, be used to mitigate the weakening of the bound in Theorem \ref{thm:main1}. In particular, combining Lemma \ref{lem:main1} with the implication of Theorem \ref{thm:main1} we obtain $\eps(\mc{D}_T, h) \leq  w_1\bar{\eps}(\mc{B}, h) + w_2\hat{\eps}(\mc{S}, h) + w_2\left(\bar{\eps}(\mc{B}, h) - \hat{\eps}(\mc{S}, h)\right)$ where $w_1 + w_2 \geq 16k$. This can be bounded by  $w_1\bar{\eps}(\mc{B}, h) + w_2\hat{\eps}(\mc{S}, h) + w_2\left(\xi(\mc{S}, \mc{B})\left\|\br_h\right\|_2 + \lambda' + R(h, \mc{S}, \mc{T})\right)$.
Therefore, it makes sense to directly optimize  $\bar{\eps}(\mc{B}, h)$ along with $ \xi(\mc{S}, \mc{B})$ and $\hat{\eps}(\mc{S}, h)$.
In this, we can assume a bound on $\left\|\br_h\right\|_2$ since the range of all $h \in \mc{F}$ is bounded in $[0,1]$. Further, the term $R(h, \mc{S}, \mc{T})$ is a difference of two unsupervised regularization terms on $\mc{S}$ and $\mc{T}$, which is expected to be small for reasonable covariate-shift in the datasets, and hence can omitted from the optimization (see Appendix \ref{app:excluding_regularization_term}). 

With this %
we formalize the above intuition to propose our loss on bags and covariate-shifted instances.\\
{\it Bags and covariate-shifted instances loss.} For parameters $\lambda_1, \lambda_2, \lambda_3 \geq 0$, the ${\sf BagCSI}$ loss is defined as:
\begin{eqnarray}
    & & {\sf BagCSI}\left(\mc{S}, \mc{B}, h, \{\lambda_i\}_{i=1}^3\right) \nonumber \\ 
    &:=& \lambda_1\bar{\eps}(\mc{B}, h) + \lambda_2\hat{\eps}(\mc{S}, h) + \lambda_3\xi^2(\mc{S}, \mc{B}) \label{eqn:BagCSI}
\end{eqnarray}
For practical considerations we use $\xi^2$ instead of $\xi$ because $\xi$ cannot be summed over mini-batches of the training dataset.

We use $\sf{BagCSI}$ loss to propose model training method in Section \ref{proposed}. We also perform extensive experiments to evaluate our methods and share the outcomes in Section \ref{sec:experiments}.
 
\input{proofs}

%% file: proofs.tex
\section{Proof of Lemma \ref{lem:main1}}\label{sec:lemmadiffbd}
Using the definitions in Section \ref{sec:prelim} define $\bu_j := (1/k)\sum_{i \in I_j} \phi(\bx_i)$ so that  $\frac{1}{k}\sum_{i \in I_j}h(\bx_i) = \br_h^{\sf T}\bu_j$. 
\begin{align}
    &\bar{\eps}(\mc{B}, h) 
    = \frac{1}{m}\sum_{j=1}^m\left[\left(\left(\frac{1}{k}\sum_{i \in I_j}h(\bx_i)\right) - y_{B_j}\right)^2\right] \nonumber \\
    = & \frac{1}{m}\sum_{j=1}^m\left[\left(\frac{1}{k}\sum_{i \in I_j}h(\bx_i)\right)^2 + y_{B_j}^2 - 2y_{B_j}\br_h^{\sf T}\bu_j \right] \nonumber \\
    \leq & \frac{1}{m}\sum_{j=1}^m\left[\frac{1}{k}\sum_{i \in I_j}(h(\bx_i)^2 + y_i^2) - 2y_{B_j}\br_h^{\sf T}\bu_j \right] \label{eqn:barepsbd}
\end{align}
where the last upper bound uses Cauchy-Schwarz inequality. On the other hand,
\begin{align}
    & \hat{\eps}(\mc{S}, h) = \frac{1}{mk}\sum_{i=1}^{mk}\left[\left(h(\bz_i) - \ell_i\right)^2\right] \nonumber \\
    = &  \frac{1}{mk}\sum_{i=1}^{mk}\left[h(\bz_i)^2 + \ell_i^2 - 2\ell_i\br_h^{\sf T}\phi(\bz_i)\right] 
\end{align}
Using the above along with \eqref{eqn:barepsbd} we obtain,
\begin{align*}
    &\bar{\eps}(\mc{B}, h) - \hat{\eps}(\mc{S}, h) \nonumber \\
    \leq & \frac{1}{mk}\sum_{i=1}^{mk}\left(h(\bx_i)^2 - h(\bz_i)^2\right) \nonumber \\ 
    & \quad + 2\br_h^{\sf T}\left(\frac{1}{m}\sum_{j=1}^{m}y_{B_j}\bu_j - \frac{1}{mk}\sum_{i=1}^{mk}\ell_i\phi(\bz_i)\right) \nonumber \\
    & \qquad + \frac{1}{mk}\sum_{i=1}^{mk}\left(y_i^2 - \ell_i^2\right)
\end{align*}
Notice that the second term on the RHS of the above is $\leq \xi(\mc{S}, \mc{B})\left\|\br_h\right\|_2$. Taking $\lambda'(\mc{S}, \mc{T})= \left|1/(mk)\sum_{i=1}^{mk}\left(y_i^2 - \ell_i^2\right)\right|$ and $R(h,\mc{S}, \mc{T}) = \left|1/(mk)\sum_{i=1}^{mk}\left(h(\bx_i)^2 - h(\bz_i)^2\right)\right|$ completes the proof of Lemma \ref{lem:main1}.

\section{Proof of Theorem \ref{thm:main1}} \label{sec:proofthmmain1}
The proof proceeds by first reformulating the process of sampling the $m$ training bags as: (i) sample $2mk$ examples from $\mc{D}_T$, (ii) partition them into $m$ disjoint $(2k)$-sized subsets, and (iii) from each subset randomly choose $k$ points to include in a bag, to obtain $m$ $k$-sized bags. First, for a fixed sample of $2mk$ examples and regressor $h \in \mc{F}$, we use the randomness in step (iii) along with concentration bounds to show that with high probability the bag-level mse loss of $h$ on the bags is at least an $O(k)$-fraction of its loss on the sampled instances. A union bound over a fine-grained $\ell_\infty$ cover of $\mc{F}$ essentially allows us to restrict ourselves to regressors in the cover. The randomness in step (i) is used along with standard generalization error bounds to show that instance-level sample loss of every $h \in \mc{F}$ can be replaced with the distributional loss. The parameter $m$ is chosen to make the error probability arbitrarily small. The rest of this section contains the formal proof.

We first describe the following equivalent way of sampling the target training bags $\mc{B} = \mc{B}(m,k) = \{(B_j,y_{B_j})\,\mid\,j=1,\dots, m\}$.
\begin{enumerate}[nolistsep,noitemsep]
    \item Let $\mc{Z} :=  \{(\bx_i, y_i)\,\mid\, i = 1,\dots, 2mk\}$ be $2mk$ iid examples from $\mc{D}_T$.
    \item Define $\ol{I}_j = \{2k(j-1)+1, \dots, 2kj\}$, $j=1, \dots, m$ be a partition of $[2mk]$ into $m$ disjoint subsets.
    \item Independently for each $j = 1,\dots, m$, let $I_j$ be a random subset of $\ol{I}_j$ of exactly $k$ indices.
    \item For each $j = 1, \dots, m$, let $B_j = \{\bx_i\,\mid\, i \in I_j\}$ with bag-labels $y_{B_j} = (1/k)\sum_{i \in I_j}y_i$.
\end{enumerate}
Let us first fix $h \in \mc{F}$ and $\mc{Z}$ and prove a lower bound on the bag-level loss.

\medskip
\noindent
\textbf{Analysis for fixed $h$ and $\mc{Z}$.}
Let us assume that $\hat{\eps}(\mc{Z}, h) = \zeta$, for some $\zeta \geq 0$. For convenience let $z_i = h(\bx_i) - y_i$, $i = 1, \dots, 2mk$. Note that since $y_i, h(\bx_i) \in [0,1]$, $|z_i| \leq 1$. Let $\ol{\mc{Z}}^{(j)} = \{(\bx_i, y_i)\,\mid\, i \in \ol{I}_j\}$ be the restriction of $\mc{Z}$ to the indices in $\ol{I}_j$, so that $\sum_{j=1}^m \hat{\eps}(\mc{Z}^{(j)}, h) = \hat{\eps}(\mc{Z}, h)$.  
Over the choice of $\{I_j\}_{j=1}^m$ define the random variable $L_j := \left[\left(\frac{1}{k}\sum_{i\in I_j}h(\bx_i)\right) - y_{B_j}\right]^2$. Since $y_{B_j} = (1/k)\sum_{i\in I_j}y_i$, $L_j = \left(\frac{1}{k}\sum_{i\in I_j}z_i\right)^2 \leq \left(\frac{1}{k}\sum_{i\in I_j}|z_i|\right)^2$.
Since $I_j \subseteq \ol{I}_j$ and $|z_i| \leq 1$ for all $i$, this implies
\begin{eqnarray}
 L_j & \leq & \tn{min}\left\{1, \left(\frac{1}{k}\sum_{i\in \ol{I}_j}|z_i|\right)^2\right\} \nonumber \\
& \leq &  \tn{min}\left\{1, \frac{2}{k}\sum_{i\in \ol{I}_j}|z_i|^2\right\} =: \gamma_j  \label{eqn:upperbd}
\end{eqnarray}
since $\sum_{i\in \ol{I}_j}|z_i| \leq \sqrt{2k}\sqrt{\sum_{i\in \ol{I}_j}|z_i|}$ by Cauchy-Schwarz inequality.
 Note that after fixing $\mc{Z}$, the choices of $I_1, \dots, I_m$ are independent of each other, and each $L_j$ only depends of the choice of $I_j$.
\begin{eqnarray}
& & \E\left[L_j\right]  =  \E\left[\left(\frac{1}{k}\sum_{i\in I_j}z_i\right)^2\right] \nonumber \\
& = & \frac{1}{k^2}\left(\sum_{r\in \ol{I}_j}z_r^2\Pr[r \in I_j] + \sum_{\substack{r,s\in \ol{I}_j\\ r\neq s}}z_rz_s\Pr[r,s\in I_j]\right) \nonumber 
\end{eqnarray}
Since $I_j$ is a random subset of $\ol{I}_j$ of $k$ out of $2k$ indices, $\Pr[r \in I_j\,\mid\, r \in \ol{I}_j] = 1/2$ and $\Pr[r, s \in I_j\,\mid\, r,s \in \ol{I}_j, r\neq s] = (k-1)/(2(2k-1))$ which simplifies the RHS of the above to:
\begin{eqnarray}
 & & \frac{1}{2k^2}\left[\left(1 - \frac{k-1}{2k-1}\right)\sum_{r\in \ol{I}_j}z_r^2 +   \frac{k-1}{2k-1}\sum_{r,s\in \ol{I}_j}z_rz_s\right] \nonumber \\
 & \geq & \frac{1}{2k^2}\left[\frac{1}{2}\sum_{r\in \ol{I}_j}z_r^2 + \frac{k-1}{2k-1}\left(\sum_{r\in \ol{I_j}}z_r\right)^2\right] \nonumber \\
 & \geq & \frac{1}{4k^2}\sum_{r\in \ol{I}_j}z_r^2 \label{eqn:explowerbd}
\end{eqnarray}
Using \eqref{eqn:upperbd} one can apply Hoeffding's inequality to obtain for any $t \geq 0$ (see Appendix \ref{app:hoeffdings}),
\begin{eqnarray}
& & \Pr\left[\sum_{j=1}^mL_j \leq \E\left[\sum_{j=1}^mL_j\right] - t\right]\nonumber \\
&\leq& 2\tn{exp}\left(\frac{-2t^2}{\sum_{j=1}^m\gamma_j^2}\right)  \nonumber \\
& \leq & 2\tn{exp}\left(\frac{-2t^2}{\left(\max\{\gamma_j\}_{j=1}^m\right)\sum_{j=1}^m\gamma_j}\right) \nonumber \\
& \leq & 2\tn{exp}\left(\frac{-t^2k}{\sum_{j=1}^m\sum_{i\in \ol{I}_j}z_i^2}\right) \nonumber
\end{eqnarray}
By definition we have $\sum_{j\in \ol{I}_j}z_i^2 = \sum_{i=1}^{2mk}z_i^2 = 2\zeta mk$. Thus, the above along with \eqref{eqn:explowerbd} yields
$\Pr\left[\sum_{j=1}^mL_j \leq \zeta m/(2k) - t\right] \leq  2\tn{exp}\left(\frac{-t^2}{2\zeta m}\right)$. Recalling that $\zeta = \hat{\eps}(\mc{Z}, h)$, and noting that $\sum_{j=1}^m L_j = m\ol{\eps}(\mc{B}, h)$ while taking $t = \zeta m /(4k)$  we obtain
\begin{equation}
    \Pr\left[\ol{\eps}(\mc{B}, h) \leq \frac{\hat{\eps}(\mc{Z}, h)}{4k}\right] \leq 2\tn{exp}\left(\frac{-\hat{\eps}(\mc{Z}, h) m}{32k^2}\right) \label{eqn:failureprob}
\end{equation}

\setlength{\tabcolsep}{3pt}

\begin{table*}[!htbp]
\captionsetup{font=small,labelfont=small}
\begin{minipage}{0.48\textwidth}
\caption{MSE scores for different methods and bag sizes on the IPUMS dataset (averaged over 10 runs). The source instance loss is $1.8714 \pm 0.08$ and target instance loss is $1.1237 \pm 0.00$. Lower is better.}
\centering
\tiny
\begin{tabular}{c|c|c|c|c}
\diagbox{{Method}}{{Bag Size}} & {8} & {32} & {128} & {256} \\ \hline
Bagged-Target & 1.14 $\pm$ 0.00 & 1.16 $\pm$ 0.00 & 1.22 $\pm$ 0.0046 & 1.31 $\pm$ 0.01 \\ 
AF & 1.23 $\pm$ 0.01 & 1.31 $\pm$ 0.01 & 1.41 $\pm$ 0.02 & 1.43 $\pm$ 0.02 \\ 
LR & 1.15 $\pm$ 0.00 & 1.18 $\pm$ 0.00 & 1.24 $\pm$ 0.01 & 1.29 $\pm$ 0.01 \\ 
AF-DANN & 1.25 $\pm$ 0.02 & 1.33 $\pm$ 0.07 & 1.39 $\pm$ 0.07 & 1.39 $\pm$ 0.02 \\
LR-DANN & 1.16 $\pm$ 0.00 & 1.23 $\pm$ 0.02 & 1.51 $\pm$ 0.07 & 1.61 $\pm$ 0.13 \\ 
DMFA & 1.15 $\pm$ 0.00 & 1.18 $\pm$ 0.00 & 1.26 $\pm$ 0.01 & 1.30 $\pm$ 0.01 \\ 
PL-WFA (our) & 1.15 $\pm$ 0.00 & 1.18 $\pm$ 0.00 & 1.25 $\pm$ 0.01 & 1.29 $\pm$ 0.01 \\ 
BL-WFA (our) & {1.14 $\pm$ 0.00} & {1.16 $\pm$ 0.00} & {1.22 $\pm$ 0.00} & {1.25 $\pm$ 0.01} \\ 
\end{tabular}
\label{tab:usc}
\end{minipage}
\hfill
\begin{minipage}{0.48\textwidth}
\caption{MSE scores for different methods and bag sizes on the Wine dataset (averaged over 20 runs). The source instance loss is $195.5 \pm 1.2$ and target instance loss is $170.5 \pm 0.1$. Lower is better.}
\centering
\tiny
\begin{tabular}{c|c|c|c|c}
\diagbox{{Method}}{{Bag Size}}& {8} & {32} & {128} & {256} \\ \hline
Bagged-Target & {173.5 $\pm$ 0.4} & {177.7 $\pm$ 1.2} & 191.0 $\pm$ 2.5 & 206.9 $\pm$ 3.5  \\ 
AF& 186.8 $\pm$ 2.1 & 190.3 $\pm$ 2.8 & 191.0 $\pm$ 2.4 & 192.4 $\pm$ 1.8  \\ 
LR& 185.9 $\pm$ 2.0 & 191.6 $\pm$ 1.6 & 193.8 $\pm$ 0.8 & 194.5 $\pm$ 1.0  \\ 
AF-DANN& 187.6 $\pm$ 1.7 & 190.5 $\pm$ 1.7 & 191.2 $\pm$ 2.5 & 191.9 $\pm$ 2.1  \\ 
LR-DANN& 186.2 $\pm$ 1.5 & 192.1 $\pm$ 2.0 & 193.7 $\pm$ 2.4 & 193.8 $\pm$ 2.5  \\ 
DMFA& 186.1 $\pm$ 1.7 & 191.8 $\pm$ 2.1 & 193.5 $\pm$ 2.4 & 194.5 $\pm$ 0.9  \\ 
PL-WFA (our)& 183.0 $\pm$ 0.6 & 186.6 $\pm$ 1.0 & 189.0 $\pm$ 0.8 & 188.9 $\pm$ 1.2  \\ 
BL-WFA (our)& 180.9 $\pm$ 0.5 & 184.6 $\pm$ 0.7 & {186.0 $\pm$ 0.8} & {186.4 $\pm$ 0.5}  \\ 
\end{tabular}
\label{tab:wine}
\end{minipage}
\end{table*}

\medskip
\textbf{High probability bound for $\mc{F}$ and $\mc{Z}$.}
Let us fix the parameter $\nu$ in the statement of Theorem \ref{thm:main1}. We fix $\mc{Z}$ for now and consider the cover $\mc{C}_\infty(\xi, \mc{F}, \mc{Z})$ for some $\xi$ which we will choose later, and $q_\infty = N_\infty(\xi, \mc{F}, 2mk)$ be the upper bound on its size. Let $\mc{C}_\tn{err} \subseteq \mc{C}_p(\xi, \mc{F}, \mc{Z})$ s.t. $\forall \hat{h} \in \mc{C}_{\tn{err}}, \hat{\eps}(\mc{Z}, \hat{h}) \geq \nu/2$.
Taking a union bound of the error in \eqref{eqn:failureprob} over $\hat{\mc{F}}_{\tn{err}}$ we obtain that:
\begin{eqnarray}
    & & \Pr\left[\forall \hat{h} \in \mc{C}_{\tn{err}}: \ol{\eps}(\mc{B}, \hat{h}) \geq \frac{\hat{\eps}(\mc{Z}, \hat{h})}{4k}\right] \nonumber \\
    & \leq& 1 - 2q_\infty\tn{exp}\left(\frac{-\nu m}{64k^2}\right) \label{eqn:unionbd}
\end{eqnarray}
Define $\hat{\mc{F}}_{\tn{err}} := \{h \in \mc{F}\,\mid\, \hat{\eps}(\mc{Z}, h) \geq 3\nu/4\}$. For any $h \in \hat{\mc{F}}_{\tn{err}}$ there is $\hat{h} \in \mc{C}_\infty(\xi, \mc{F}, \mc{Z})$ s.t. $|\hat{h}(\bx) - h(\bx)| \leq \xi$ for all $(\bx, y) \in \mc{Z}$. Now, $(\hat{h}(\bx) - y)^2 = (h(\bx) - y + \hat{h}(\bx) - h(\bx))^2 \geq (h(\bx) - y)^2 - 2|\hat{h}(\bx) - h(\bx)||h(\bx) - y| + (\hat{h}(\bx) - h(\bx))^2 \geq (h(\bx) - y)^2 - 2\xi$ since $h(\bx), y \in [0,1]$.
Similarly, consider any bag $B \in \mc{B}$. Using arguments analogous to above we obtain $\left(\E\left[h(\bx)\right] - y_B\right)^2 \geq \left(\E\left[\hat{h}(\bx)\right] - y_B\right)^2 - 2\left|\E[\hat{h}(\bx) - h(\bx)]\right|\left(\E\left[h(\bx)\right] - y_B\right) \geq \left(\E\left[\hat{h}(\bx)\right] - y_B\right)^2 - 2\xi$, implying
\begin{equation}
    \hat{\eps}(\mc{Z}, \hat{h}) \geq \hat{\eps}(\mc{Z}, h) - 2\xi, \quad \ol{\eps}(\mc{B}, h) \geq \ol{\eps}(\mc{B}, \hat{h}) - 2\xi. \label{eqn:hatbd}
\end{equation}
Therefore, taking $\xi = \nu/(32k)$ we obtain from the first bound above that $\hat{h} \in \mc{C}_{\tn{err}}$ and further that $\hat{\eps}(\mc{Z}, \hat{h}) \geq 2\hat{\eps}(\mc{Z}, h)/3 \geq \nu/2 = 16k\xi$. Observe that $\ol{\eps}(\mc{B}, \hat{h}) \geq \hat{\eps}(\mc{Z}, \hat{h})/(4k)$ implies $\ol{\eps}(\mc{B}, \hat{h}) \geq 4\xi$, which in turn implies  $\ol{\eps}(\mc{B}, h) \geq \ol{\eps}(\mc{B}, \hat{h}) - 2\xi \geq \ol{\eps}(\mc{B}, \hat{h})/2$. Combining this with \eqref{eqn:unionbd} and \eqref{eqn:hatbd} we obtain,
\begin{align}
    \Pr&\left[\forall h \in \hat{\mc{F}}_{\tn{err}}: \ol{\eps}(\mc{B}, h) \geq \frac{\hat{\eps}(\mc{Z}, h)}{12k}\right] \nonumber \\
    &\leq 1 - 2q_\infty\tn{exp}\left(\frac{-\nu m}{64k^2}\right) \label{eqn:unionbd-2}
\end{align}
We now unfix $\mc{Z}$, and define $\mc{F}_{\tn{err}} = \{h \in \mc{F}\,\mid\, {\eps}(\mc{D}_T, \hat{h}) \geq \nu\}$. By Theorem 17.1 of \cite{Anthony-Bartlett}, we obtain with probability at least $1 - 4q_1\tn{exp}\left(-2\nu^2mk/512\right)$ over the choice of $\mc{Z}$, $h \in \mc{F}_\tn{err} \Rightarrow h \in \hat{\mc{F}}_{\tn{err}}$ where $q_1 = N_1(\nu/64, \mc{F}, 4mk)$. Using this along with \eqref{eqn:unionbd-2}, we obtain that with probability at least $1 - 2q_\infty\tn{exp}\left(-\nu m/(64k^2)\right) - 4q_1\tn{exp}\left(-2\nu^2mk/512\right)$, $\forall h  \in \mc{F}_{\tn{err}}, \ol{\eps}(\mc{B}, h) \geq \frac{\hat{\eps}(\mc{Z}, h)}{12k} \geq \frac{3\nu}{48 k} = \frac{\nu}{16 k}$. Using the upper bounds in \eqref{eqn:coversize} we see that the probability is $1 - \delta$ if we choose $m \geq
O\left(\left(p\left(\log\left(\frac{k}{\nu}\right) + \log\log\left(\frac{1}{\delta}\right)\right) + \log\frac{1}{\delta}\right)\max\left\{\frac{1}{k\nu^2}, \frac{k^2}{\nu}\right\}\right)$. See Appendix \ref{app:sample_complexity_analysis} for more details.
This completes the proof of Theorem \ref{thm:main1}. %

%% file: proposed.tex
\section{Proposed Methods}\label{proposed}

We propose two novel methods. The first method uses ${\sf BagCSI}$ loss as the objective. We have shown above that ${\sf BagCSI}$ loss is an upper bound over $\eps(\mc{D}_T, h)$ loss w.r.t target distribution. We now provide intuitive explanation for why ${\sf BagCSI}$ loss should work.

Let us assume that the goal is to predict label for an unseen instance $\bx$, given feature representations $\phi(\bx_i)$ in the embedding space and corresponding labels $y_i$ from training data.
A natural prediction would be $\E_i \left[\rho(\phi(\bx), \phi(\bx_i))y_i\right]$, where $\rho$ is some similarity metric.
If we choose the similarity metric to be the inner product, the prediction can be written as $\phi(\bx)^{\sf T}\E_i \left[\phi(\bx_i)y_i\right]$.
The given feature representations and corresponding labels can come either from the source domain or from the target domain. 
For learning domain invariant feature representation, the prediction should be similar irrespective of the domain considered. 
This can be achieved by enforcing the term, $\sum\limits_i y_i\phi(\bx_i)$ to be equal for source and target domain. However, this approach requires knowledge of instance-level labels $y_\bx$ from target domain, which are not available. We can however replace $y_\bx$ with \emph{pseudo-labels} $\hat{y}_\bx$, using which we introduce a new domain adaptation loss term in the objective, $\psi^2(\mc{S}, \mc{B})$ where: 
\begin{align}
 \psi(\mc{S}, \mc{B}) & := \frac{1}{mk}\left\|\sum_{j=1}^m \sum_{\bx \in B_j}\hat{y}_\bx\phi(\bx) - \sum_{i=1}^{mk}y_i\phi(\bz_i) \right\|_2 \label{eq:domain_adaptation_loss}
\end{align}
One way is to assign the bag-label as the pseudo-label for all instances withing the bag, in which case $\psi(\mc{S}, \mc{B})$ essentially reduces to $\xi(\mc{S}, \mc{B})$. %
We call this method \textit{Bag Label Weighted Feature Alignment} (\textbfne{BL-WFA}) which involves training using the ${\sf BagCSI}$ loss.

Another approach is to use the following process for pseudo-labeling instances in a bag $B$ using hypothesis model $h$:
\begin{enumerate}[nolistsep,noitemsep,leftmargin=*]
    \item Compute the predictions $\{h(\bx)\}_{\bx\in B}$.
    \item The pseudo-labels are given by adding to each prediction the same $b \in \R$ such that average pseudo-label in the bag equals the bag-label. Note that this  is equivalent to  the nearest  vector of pseudo-labels  (in Euclidean distance) to the vector predictions, that satisfies the bag-label constraint.
\end{enumerate}
We call this method \textit{Pseudo-label Weighted Feature Alignment} (\textbfne{PL-WFA}) in which  $\psi(\mc{S}, \mc{B})$ is used to train the model using the above computed pseudo-labels.

%% file: experiments.tex
\section{Experimental Evaluations}\label{sec:experiments}

\begin{table*}[!htbp]
\captionsetup{font=small,labelfont=small}
\centering
\tiny
\begin{minipage}{0.48\textwidth}
\caption{MSE scores for different methods and bag sizes on the Synthetic dataset (averaged over 20 runs). The source instance loss is $2718.13 \pm 2062.32$ and target instance loss is $0.19 \pm 0.02$. Lower is better.}
\begin{tabular}{c|c|c|c|c}
\diagbox{{Method}}{{Bag Size}}& {8} & {32} & {128} & {256} \\ \hline
Bagged-Target& 0.71 $\pm$ 0.05 & 5.49 $\pm$ 0.93 & 17.87 $\pm$ 0.49 & 19.95 $\pm$ 0.34  \\ 
AF& 0.96 $\pm$ 0.07 & 6.22 $\pm$ 0.81 & 18.16 $\pm$ 0.50 & 20.00 $\pm$ 0.86  \\ 
LR& 0.71 $\pm$ 0.04 & 5.15 $\pm$ 1.06 & 18.10 $\pm$ 0.40 & 19.92 $\pm$ 1.55  \\ 
AF-DANN& 1.23 $\pm$ 0.06 & 8.16 $\pm$ 0.54 & 18.04 $\pm$ 0.95 & 20.15 $\pm$ 0.49  \\ 
LR-DANN& 1.02 $\pm$ 0.04 & 7.84 $\pm$ 0.87 & 17.76 $\pm$ 0.24 & 19.72 $\pm$ 0.29  \\ 
DMFA& {0.69 $\pm$ 0.05} & 4.39 $\pm$ 0.84 & 16.50 $\pm$ 1.47 & 19.07 $\pm$ 1.16  \\ 
PL-WFA (our)& 0.75 $\pm$ 0.06 & 4.43 $\pm$ 0.81 & 15.60 $\pm$ 0.94 & 18.40 $\pm$ 0.74  \\ 
BL-WFA (our)& 0.75 $\pm$ 0.05 & {2.22 $\pm$ 0.22} & {10.36 $\pm$ 3.15} & {13.76 $\pm$ 0.60}  \\ 
\end{tabular}
\label{tab:synthetic}
\end{minipage}
\hfill
\begin{minipage}{0.48\textwidth}
\caption{MSE scores for different methods and bag sizes on the Criteo SSCL dataset (averaged over 10 runs). The source instance loss is $293.74 \pm 5.1$ and target instance loss is $147.79 \pm 0.3$. Lower is better.}
\begin{tabular}{c|c|c|c|c}
\diagbox{{Method}}{{Bag Size}} & {64} & {128} & {256} & {512} \\ \hline
Bagged-Target &208.78 $\pm$ 2.7 &234.32 $\pm$ 3.3 &254.78 $\pm$ 5.3 &264.74 $\pm$ 5.3 \\
AF &297.95 $\pm$ 6.5 &296.51 $\pm$ 6.1 &294.86 $\pm$ 5.3 &299.93 $\pm$ 6.5 \\
LR &207.78 $\pm$ 2.7 &232.72 $\pm$ 10. &256.68 $\pm$ 13. &264.46 $\pm$ 5.4 \\
AF-DANN &296.95 $\pm$ 6.3 &296.35 $\pm$ 6.4 &295.49 $\pm$ 5.2 &297.91 $\pm$ 7.3 \\
LR-DANN &206.39 $\pm$ 2.3 &230.84 $\pm$ 3.1 &243.62 $\pm$ 4.5 &265.33 $\pm$ 4.6 \\
DMFA &207.60 $\pm$ 2.7 &232.40 $\pm$ 9.9 &247.66 $\pm$ 3.4 &264.51 $\pm$ 5.5 \\
PL-WFA (our) &204.71 $\pm$ 2.6 &226.39 $\pm$ 2.9 &240.55 $\pm$ 3.3 &254.46 $\pm$ 5.5 \\
BL-WFA (our) &{204.62 $\pm$ 2.4} &{226.33 $\pm$ 2.9} &{240.39 $\pm$ 3.2} &{254.36 $\pm$ 5.5} \\

\end{tabular}
\label{tab:criteo}
\end{minipage}
\end{table*}

We evaluate our approaches via experiments on both synthetic as well as real-world datasets and compare against the baselines for different bag sizes.

\medskip
\noindent
{\bf Baseline Methodologies.}
In \cite{Li-Culotta}, authors propose methods for domain adaptation in LLP setting for classification tasks. We adapt these methods for regression tasks and consider those as baselines. In this paper, these baselines are referred to as Average Feature ({AF}), Label Regularization ({LR}), Average Feature DANN ({AF-DANN}) and Label Regularization DANN ({LR-DANN}). See Sections 3.1.2, 3.1.3, 3.2.1, 3.2.2 in \cite{Li-Culotta} for respective methods.
In literature on domain adaptation (for non-LLP settings) \citep{long2015learning, long2017deep}, it has been shown that approaches using MMD (maximum mean discrepancy) based objectives work well. Hence, we also define a baseline that uses similar objective adapted for our setting, called Domain Mean Feature Alignment ({DMFA}).
We also consider bag level target loss ({Bagged-Target}) as a baseline. Appendix \ref{app:baselines} contains additional details about baseline methods.
We evaluate and compare our methods against these baselines.

Our model training uses the above losses in a mini-batch loop.
For DMFA and PL-WFA we select equal number of instances from both source and target domain in a mini-batch. 
For BL-WFA, we select as many instances from source domain as the number of bags selected from target domain in a mini-batch.
Such a choice avoids explicit normalization in the objective function and incorporates them into the hyper-parameters.
 We evaluate all the baselines and proposed methods for different bag sizes and datasets. 

\medskip
\noindent
{\bf Synthetic Dataset.} The synthetic dataset has 64 dimensional continuous feature vectors and scalar-valued continuous label. For covariate shifted source and target domain data, the feature vectors are sampled from a multi-dimensional Gaussian distribution with different means and covariance matrices. The labels for both source and target data are computed using the same randomly initialized neural network. We also perform ablation studies to observe the impact of magnitude of covariance shift.
The train set comprises 0.2 million instances from both source and target domain. The test set comprises 65 thousand instances from target domain.

\input{tab_correlated_bags}

\begin{table}[htbp]
\captionsetup{font=small,labelfont=small}
\begin{minipage}{0.48\textwidth}
\centering
\caption{MSE scores on Criteo dataset with correlated bags. Lower is better.}\label{tab:correlated_criteo}
\tiny
\begin{tabular}{l|r|r|r|r}
\diagbox{{Method}}{{Bag Size}}&64 &128 &256 &512 \\\midrule
Bagged-Target &204.78 $\pm$ 2.7 &211.12 $\pm$ 3.1 &226.78 $\pm$ 3.8 &254.74 $\pm$ 5.3 \\
AF &257.92 $\pm$ 2.0 &266.94 $\pm$ 3.4 &276.82 $\pm$ 3.1 &294.13 $\pm$ 4.4 \\
LR &179.88 $\pm$ 0.6 &183.77 $\pm$ 1.1 &191.25 $\pm$ 1.1 &207.24 $\pm$ 1.2 \\
AF-DANN &257.48 $\pm$ 2.1 &263.87 $\pm$ 0.6 &275.14 $\pm$ 3.5 &292.43 $\pm$ 5.0 \\
LR-DANN &179.37 $\pm$ 0.5 &183.73 $\pm$ 1.0 &191.17 $\pm$ 1.1 &207.98 $\pm$ 1.6 \\
DMFA &180.89 $\pm$ 0.6 &183.47 $\pm$ 1.2 &191.27 $\pm$ 1.2 &207.18 $\pm$ 1.2 \\
PL-WFA (our) &177.76 $\pm$ 0.7 &181.70 $\pm$ 1.2 &188.29 $\pm$ 1.2 &197.23 $\pm$ 1.2 \\
BL-WFA (our) &{177.74 $\pm$ 0.7} &{181.66 $\pm$ 1.1} &{188.19 $\pm$ 1.2} &{197.07 $\pm$ 1.3} \\
\end{tabular}
\end{minipage}
\end{table}

\noindent
{\bf Real-world Datasets.} We also evaluate methods on three real world datasets: \textit{Wine Ratings}~\citep{wine_ratings, wine_reviews}, \textit{IPUMS USA}~\citep{ipums_usa_2024} Census data, and \textit{Criteo Sponsored Search Conversion Logs (SSCL)}~\citep{tallis2018reacting}.

{\it Wine}: We use Price column as the label. The source domain comprises of wines from France and the target domain comprises of wines from all countries but France.
The train set comprises 0.5 million instances from both source and target domain. The test set comprises 0.2 million instances from target domain.
\\
{\it IPUMS USA}: We use INCWAGE column as the label. We consider the data from 1970 as the source domain and data from 2022 as the target domain.
The train set comprises 1.3 million instances from source and 9.4 million instances from target domain. The test set comprises 0.3 million instances from target domain. 
\\
{\it Criteo SSCL}: We use SalesAmountInEuro column as the label. We create a domain split on the basis of the country field (the most frequently occurring country in the dataset as source and rest as the target).
The train set comprises 0.5 million instances from source and 0.9 million instances from target domain. The test set comprises 0.2 million instances from target domain.

Appendix \ref{app:dataset} contains details about size and pre-processing for all the datasets.

All datasets are split into two components, source and target domain. For our study, it is important that there is a reasonable covariate shift between these two components. The target domain dataset is split into train (80\%) and test (20\%) sets.
The target domain component of train set is partitioned randomly into bags of equal size.
We also perform experiments with correlated bags. To partition the dataset into correlated bags, we select a feature and create bags such that all the samples in that bag have the same value of that feature if the feature is categorical. If the feature is numerical, we sort the dataset on the basis of that feature and use consecutive samples for creating the bags. Further details about creation of correlated bags are provided in Appendix \ref{app:correlated_bags}.
Additionally, we also perform experiments by partitioning the dataset into bags of mixed (non-uniform) sizes. We do so in two different ways; SBB (Sample Balanced Bagging - equal number of instances for each bag size) and BBB (Bag Balanced Bagging - equal number of bags of each size). Each bag in the resultant dataset is of the size 8, 32, 128 or 256. Further details about partitioning the dataset into mixed bag sizes are provided in Appendix \ref{app:mixed_bag_sizes}.

\medskip
\noindent
{\bf Training \& Evaluation.}
We use a simple neural network comprising of an input layer followed by two sequential ReLU activated layers (128 nodes) and a final linear layer (1 node). For IPUMS and Criteo SSCL datasets, we additionally include embedding layers after the input layer for all the cardinal and categorical features that were not converted to one-hot representations. For AF-DANN and LR-DANN, we also have a sigmoid activated domain prediction layer in parallel to the final dense layer.

During training, we perform a grid search to find the most optimal set of hyperparameters for each configuration (specific dataset, methodology and bag size). We try out two different optimizers for all experiments mentioned in the main paper - Adam and SGD and report scores corresponding to the best performer. We observed that Adam works better for most of the cases, so we perform experiments described in Appendix with Adam optimizer only. See Appendix \ref{app:hyperparameters} for more details.

For each configuration, we run the same experiment multiple times and report the MSE scores on target domain's test data as the evaluation metric. Note that the instances in target domain are randomly bagged for each run. The final evaluation metric is reported by the mean and standard deviation over these runs. We run 20 trials for each configuration with Wine and Synthetic datasets and 10 trials for each configuration with IPUMS and Criteo SSCL datasets.

{\bf Experimental Code and Resources.}\footnote{The code for our experiments can be found at \url{www.github.com/google-deepmind/covariate_shifted_llp}.} Our experiments were run on a system with standard 8-core CPU, 256GB of memory with one P100 GPU.

MSE scores on IPUMS, Wine, Synthetic and Criteo SSCL datasets with random bagging for different bag sizes are reported in Tables \ref{tab:usc}, \ref{tab:wine}, \ref{tab:synthetic} and \ref{tab:criteo}.
MSE scores on Wine, Criteo SSCL and IPUMS datasets with random bagging for BBB-mixed and SBB-mixed bag sizes are reported in Tables \ref{tab:bbb_mixed_bag_size} and \ref{tab:sbb_mixed_bag_size}.
MSE scores with correlated bags for IPUMS, Synthetic and Criteo datasets  are reported in Tables \ref{tab:correlated_ipums}, \ref{tab:correlated_synthetic} and \ref{tab:correlated_criteo}  respectively.

Results for more experiments are reported in Appendix \ref{app:results}. This includes experiments on Wine dataset with a different domain split (see Table \ref{tab:wine_italy}), experiments on synthetic dataset with a non-diagonal covariance matrix (see Table \ref{tab:synthetic_non_diagonal}), and experiments on synthetic dataset by varying the magnitude of covariate shift (see Table \ref{tab:synth-perturbation}).

\input{tab_mix_bags}

%% file: tab_correlated_bags.tex
\begin{table*}[htbp]
\begin{minipage}{0.48\textwidth}
\captionsetup{font=small,labelfont=small}
\caption{MSE scores on IPUMS dataset with correlated bags. Lower is better.}\label{tab:correlated_ipums}
\tiny
\begin{tabular}{l|r|r|r|r}
\diagbox{{Method}}{{Bag Size}}&8 &32 &128 &256 \\\midrule
Bagged Target &1.09 $\pm$ 0.00 &1.12 $\pm$ 0.00 &1.17 $\pm$ 0.00 &1.24 $\pm$ 0.00 \\
AF &1.31 $\pm$ 0.00 &1.35 $\pm$ 0.01 &1.38 $\pm$ 0.01 &1.41 $\pm$ 0.01 \\
LR &1.10 $\pm$ 0.00 &1.12 $\pm$ 0.00 &1.18 $\pm$ 0.00 &1.25 $\pm$ 0.00 \\
AFDANN &1.31 $\pm$ 0.01 &1.35 $\pm$ 0.01 &1.37 $\pm$ 0.01 &1.38 $\pm$ 0.01 \\
LRDANN &1.09 $\pm$ 0.00 &1.11 $\pm$ 0.00 &1.17 $\pm$ 0.00 &{1.21 $\pm$ 0.00} \\
DMFA &1.10 $\pm$ 0.00 &1.12 $\pm$ 0.00 &1.18 $\pm$ 0.00 &1.22 $\pm$ 0.00 \\
PLWFA &{1.08 $\pm$ 0.00} &{1.10 $\pm$ 0.00} &{1.16 $\pm$ 0.00} &{1.21 $\pm$ 0.00} \\
BLWFA &{1.08 $\pm$ 0.00} &{1.10 $\pm$ 0.00} &{1.16 $\pm$ 0.00} &{1.21 $\pm$ 0.00} \\
\end{tabular}
\end{minipage}
\hfill
\begin{minipage}{0.48\textwidth}
\captionsetup{font=small,labelfont=small}
\caption{MSE scores on Synthetic dataset with correlated bags. Lower is better.}\label{tab:correlated_synthetic}
\tiny
\begin{tabular}{l|r|r|r|r}
\diagbox{{Method}}{{Bag Size}}&8 &32 &128 &256 \\\midrule
Bagged-Target &0.65 $\pm$ 0.07 &2.13 $\pm$ 0.35 &7.01 $\pm$ 0.61 &10.35 $\pm$ 1.33 \\
AF &1.02 $\pm$ 0.17 &4.20 $\pm$ 0.52 &9.65 $\pm$ 0.54 &12.35 $\pm$ 0.89 \\
LR &0.60 $\pm$ 0.05 &2.01 $\pm$ 0.33 &6.72 $\pm$ 0.78 &9.18 $\pm$ 0.74 \\
AF-DANN &1.30 $\pm$ 0.16 &5.14 $\pm$ 0.46 &10.67 $\pm$ 0.43 &13.42 $\pm$ 0.88 \\
LR-DANN &0.81 $\pm$ 0.08 &3.26 $\pm$ 0.38 &8.31 $\pm$ 0.63 &10.37 $\pm$ 0.78 \\
DMFA &0.57 $\pm$ 0.05 &2.17 $\pm$ 0.28 &6.35 $\pm$ 0.56 &8.86 $\pm$ 0.69 \\
PL-WFA (our) &{0.56 $\pm$ 0.04} &{1.90 $\pm$ 0.30} &{6.09 $\pm$ 0.69} &8.50 $\pm$ 0.60 \\
BL-WFA (our) &0.60 $\pm$ 0.05 &1.90 $\pm$ 0.31 &6.11 $\pm$ 0.74 &{8.48 $\pm$ 0.81} \\
\end{tabular}
\end{minipage}
\end{table*}

%% file: tab_mix_bags.tex
\begin{table*}[htbp]
\begin{minipage}{0.48\textwidth}
\captionsetup{font=small,labelfont=small}
\caption{MSE scores for data with BBB mixed bag sizes. BBB is bag balanced bagging i.e. there are equal number of bags of each size. Lower is better.}\label{tab:bbb_mixed_bag_size}
\tiny
\begin{tabular}{l|rr|rr|rr}
\diagbox{{Method}}{{Dataset}} & &Wine & &Criteo & &IPUMS \\\midrule
Bagged Target & &219.51 $\pm$ 7.96 & &209.37 $\pm$ 1.98 & &1.26 $\pm$ 0.01 \\
AF & &210.45 $\pm$ 6.58 & &210.24 $\pm$ 2.54 & &1.27 $\pm$ 0.01 \\
LR & &211.58 $\pm$ 7.67 & &213.01 $\pm$ 3.01 & &1.25 $\pm$ 0.01 \\
AFDANN & &289.16 $\pm$ 2.69 & &205.36 $\pm$ 2.65 & &1.26 $\pm$ 0.01 \\
LRDANN & &190.88 $\pm$ 1.34 & &208.03 $\pm$ 2.79 & &1.28 $\pm$ 0.01 \\
DMFA & &193.29 $\pm$ 3.90 & &207.80 $\pm$ 3.14 & &1.26 $\pm$ 0.01 \\
PLWFA & &183.42 $\pm$ 3.76 & &202.94 $\pm$ 2.96 & &1.25 $\pm$ 0.01 \\
BLWFA & &{183.00 $\pm$ 3.13} & &{202.75 $\pm$ 2.86} & &{1.24 $\pm$ 0.01} \\
\end{tabular}
\end{minipage}
\hfill
\begin{minipage}{0.48\textwidth}
\captionsetup{font=small,labelfont=small}
\caption{MSE scores for data with SBB mixed bag sizes. SBB is sample balanced bagging i.e. for a particular bag size, there are equal number of samples. Lower is better.}\label{tab:sbb_mixed_bag_size}
\tiny
\begin{tabular}{l|rr|rr|rr}
\diagbox{{Method}}{{Dataset}} & &Wine & &Criteo & &IPUMS \\\midrule
Bagged Target & &183.94 $\pm$ 2.00 & &177.69 $\pm$ 0.85 & &1.16 $\pm$ 0.01 \\
AF & &186.04 $\pm$ 1.43 & &176.99 $\pm$ 0.82 & &1.15 $\pm$ 0.01 \\
LR & &186.07 $\pm$ 1.46 & &181.89 $\pm$ 1.33 & &1.15 $\pm$ 0.01 \\
AFDANN & &163.51 $\pm$ 0.36 & &177.13 $\pm$ 0.45 & &1.17 $\pm$ 0.01 \\
LRDANN & &163.52 $\pm$ 0.37 & &181.22 $\pm$ 0.99 & &1.18 $\pm$ 0.01 \\
DMFA & &163.29 $\pm$ 1.08 & &181.62 $\pm$ 1.25 & &1.16 $\pm$ 0.01 \\
PLWFA & &{161.31 $\pm$ 1.03} & &171.84 $\pm$ 1.34 & &1.16 $\pm$ 0.01 \\
BLWFA & &161.59 $\pm$ 1.11 & &{171.71 $\pm$ 1.31} & &{1.14 $\pm$ 0.01} \\
\end{tabular}
\end{minipage}
\end{table*}

%% file: results.tex
\section{Results \& Inferences}
The evaluation metrics are reported for all 3 datasets in tables \ref{tab:wine}, \ref{tab:usc} and \ref{tab:synthetic}. For bag size 256, BL-WFA achieves 2.9\%, 2.5\% and 27.9\% improvement over the best baseline method for Wine, US Census and synthetic datasets respectively. MSE is used as evaluation metric, hence scores cannot be compared across datasets due to different scales. Refer Appendix \ref{app:results} for more results.

We make the following inferences from the results:
\begin{enumerate}
    \item PL-WFA and BL-WFA consistently outperform all other baselines for large enough bag sizes. This is expected because with increase in bag size, the information from just the bagged target domain is not rich enough and benefits greatly from inclusion of covariate shifted source domain data. By leveraging not just the features from target domain but also the bagged-labels, PL-WFA and BL-WFA outperform other baseline methods which rely only on features from target domain for domain adaptation.
    \item With increase in bag size, the performance drops. This is expected as information is lost with increase in bag size.
    \item On synthetic dataset (where we definitely have a reasonable amount of covariate shift), even with bag size as large as 256, we see that the performance of our proposed methods - PL-WFA and BL-WFA is better than the case where we use instance level labeled target data for training ( target instance loss). This improvement is achieved despite the fact that performance when just using the source data for training (source instance loss) is poor.
    \item On smaller bag sizes, other methods (for example, LR and DMFA on synthetic dataset and Bagged-Target on Wine dataset) seem to outperform our proposed methods. Such behavior is expected when the information from target data is itself sufficient to learn a good enough function approximator. It is worth noting that the objective function in our proposed method reduces to that of LR for $\lambda = 0$. So, in theory PL-WFA and BL-WFA are always better than LR. By decreasing the $\lambda$ value, our methods can do at least as good as LR.
    \item Although the best baseline method is different for different datasets under consideration (AF-DANN on Wine, LR on US Census and DMFA on synthetic), BL-WFA consistently beats the best baseline for a large enough bag size.
\end{enumerate}

%% file: conclusion.tex
\section{Conclusion}
We formally define the problem of learning from label aggregates where source data has instance wise labels while target data has aggregate labels of instances grouped into bags. We also give bag-to-instance generalization error bound for regression tasks in LLP and use it to arrive at BagCSI loss. We propose two new methods, BL-WFA (based on BagCSI) and PL-WFA (based on a variant of BagCSI) that naturally incorporate the knowledge of aggregate labels from target domain in the domain adaptation framework leading to improvement over baseline methods. We also adapt several methods from literature in domain adaptation and LLP to this setting. Through experiments on synthetic and real-world datasets we show that our methods consistenty outperform baseline techniques.

%% file: supplement.tex
\onecolumn
\section{Useful Concepts}\label{app:simplifying}
\subsection{Embedding Space Representation}
 For a Hilbert space $\mc{H}$ of real-valued functions defined over $\mbc{X}$, for every $\bx \in \mbc{X}$ s.t. the mapping $L_\bx : \mc{H} \to \R$ given by $L_\bx(f) = f(\bx)$ is bounded i.e., $|L_\bx(f)| \leq C_\bx \|f\|_\mc{H}$, the Riesz Representation Theorem guarantees the existence of $g_\bx \in \mc{H}$ s.t. $L_\bx(f) = \langle f, g_\bx\rangle_\mc{H}$. As we study regression tasks (typically neural regression) in this work, we can assume boundedness and define $f(\bx) = \br_f^{\sf T}\phi(\bx)$ where $\phi$ is a mapping to a real-vector in an embedding space, and $\br_f$ the representation of $f$ in that space.

The function class under consideration in our experiments is a neural network with the final layer being a single node (without any activation) as we are studying the scalar regression use-case. In this case, the embedding space is learnt during training.  Here, $\phi(\bx)$ is the output of penultimate layer of neural network and $\br_f$ are the parameters of the final layer (a single node). 

\subsection{Excluding Regularization Term in Loss Function}\label{app:excluding_regularization_term}
The regularization term $R(h,\mc{S}, \mc{T}) = \left|1/(mk)\sum_{i=1}^{mk}\left(h(\bx_i)^2 - h(\bz_i)^2\right)\right|$ enforces that the \textit{average} squared-predictions of $h$ i.e. 
the squared $\ell_2$-norm of $h$, on the source and the target domains should be similar. However, covariate-shifts often approximately preserve the $\ell_2$-norm of predictors for e.g. if they are \emph{rotational} in the embedding space $\{\phi(\bx)\}$. Therefore, for practical settings %
the contribution of $R(h,\mc{S}, \mc{T})$ (for example, to gradient updates in neural networks) can be ignored and the term is omitted  from the ${\sf BagCSI}$ loss. 

This claim is empirically validated in Tables \ref{tab:reg_term_mag_ipums}, \ref{tab:reg_term_mag_criteo} and \ref{tab:reg_term_mag_wine} which establish that the magnitude of $R(h,\mc{S}, \mc{T})$ term is very small compared to ${\sf BagCSI}$. We report the average loss values over 5 random partitionings of the training data into bags.

It is also established empirically that adding the regularization term $R(h,\mc{S}, \mc{T})$ in the loss does not result in significant improvement. This can be observed in the experimental results presented in the Tables \ref{tab:reg_term_train_ipums}, \ref{tab:reg_term_train_criteo} and \ref{tab:reg_term_train_wine} which are obtained by doing a hyperparameter search within a range $W = \{10^{-5}, 5\times 10^{-5}, 10^{-4}, 5\times 10^{-4}, 10^{-3}, 5\times 10^{-3}, 10\times {-2}\}$ of the weight for the regularization term in the overall loss.

\subsection{Sample Complexity Analysis}\label{app:sample_complexity_analysis}
Given that with probability at least $1 - 2q_\infty\tn{exp}\left(-\nu m/(64k^2)\right) - 4q_1\tn{exp}\left(-2\nu^2mk/512\right)$, $\forall h  \in \mc{F}_{\tn{err}}, \ol{\eps}(\mc{B}, h) \geq  \frac{\nu}{16 k}$, we show that if we chose $m \geq O\left(\left(p\left(\log\left(\frac{k}{\nu}\right) + \log\log\left(\frac{1}{\delta}\right)\right) + \log\frac{1}{\delta}\right)\max\left\{\frac{1}{k\nu^2}, \frac{k^2}{\nu}\right\}\right)$, then with probability at least $1-\delta$, $\forall h  \in \mc{F}_{\tn{err}}, \ol{\eps}(\mc{B}, h) \geq  \frac{\nu}{16 k}$.

Note that,
$q_1 = N_1(\nu/64, \mc{F}, 4mk)$ and
$q_\infty = N_\infty(\nu/32k, \mc{F}, 2mk)$.

From (\ref{eqn:coversize}), $N_1(\xi, \mc{F}, N) \leq N_\infty(\xi, \mc{F}, N) \leq (eN/\xi p)^p$. Hence, $q_1 \leq \left(\frac{256emk}{\nu p}\right)^p$ and $q_\infty \leq \left(\frac{64emk^2}{\nu p}\right)^p$.

Let $R_\infty = q_\infty\tn{exp}\left(-\nu m/(64k^2)\right)$ and $R_1 = q_1\tn{exp}\left(-2\nu^2mk/512\right)$.

Substituting $m = c\left(p\left(log\left(\frac{k}{\nu}\right) + \log\log\left(\frac{1}{\delta}\right)\right) + log\frac{1}{\delta}\right)\max\left\{\frac{1}{k\nu^2}, \frac{k^2}{\nu}\right\}$, where $c$ is some large constant,

\begin{equation*}
    \log R_\infty = p\log\left(\frac{64emk^2}{\nu p}\right) - \frac{\nu m}{64k^2} \\
    \leq p \left(\log64em + \log\frac{k^2}{\nu} - \log p\right) - \frac{c}{64}\left(p\log\frac{k}{\epsilon} + p\log\log\frac{1}{\delta} + \log\frac{1}{\delta}\right).
\end{equation*}

As $\log64em \leq \log64ec + \log p + \log\log\frac{k}{\nu} + \log\log\log\frac{1}{\delta} + \log\log\frac{1}{\delta} + \log\frac{k^2}{\nu} + \log\frac{1}{k\nu^2}$,

\begin{align*}
    \log R_\infty &\leq p \left[\log64ec + \log p + \log\log\frac{k}{\nu} + \log\log\log\frac{1}{\delta} + \log\log\frac{1}{\delta} + \log\frac{k^2}{\nu} + \log\frac{1}{k\nu^2} \log\frac{k^2}{\nu} - \log p\right] \\
    & \quad\quad\quad\quad- \frac{c}{64}\left(p\log\frac{k}{\nu} + p\log\log\frac{1}{\delta} + \log\frac{1}{\delta}\right) \\
    &\leq -\log\left(\frac{4}{\delta}\right),
\end{align*}
for a large enough constant $c$ and for small enough $\delta$.
Hence, $R_\infty \leq \delta/4$. Using a similar analysis, we also obtain that, $R_1 \leq \delta/8$. Thus, $1-2R_\infty-4R_1 \geq 1-\delta$ follows for a large enough constant $c$ and small enough $\delta$, completing the proof.

\section{Useful Analytical Tools}
\subsection{Hoeffding's Inequality}\label{app:hoeffdings}
We use the Hoeffding's inequality which is stated below.
\begin{theorem}[Hoeffding]\label{thm:hoeffding}
 Let $X_1,\dots, X_n$ be independent random variables, s.t. $a_i \leq X_i \leq b_i$, $\Delta_i = b_i - a_i$ for $i = 1,\dots, n$. Then, for any $t > 0$,
	$$\Pr\left[\left|\sum_{i=1}^n X_i - \sum_{i=1}^n\E[X_i]\right| > t\right] \leq 2\cdot\tn{exp}\left(-\frac{2t^2}{\sum_{i=1}^n\Delta_i^2}\right).$$
\end{theorem}

\subsection{Pseudo-Dimension}\label{app:pseudo-dimension}
As defined in Section \ref{sec:prelim}, $\mc{F}$ is a class of real-values functions (regressors) mapping $\mathbb{R}^d$ to $[0,1]$.

A finite subset $\mbc{X} = \{x_1, x_2, \ldots, x_N\} \subset \mathbb{R}^d$ is \textit{pseudo-shattered} by \mc{F} if there exist $r_1, r_2,\ldots, r_N$ such that for each $b \in \{0, 1\}^m$, there is a function $f_b$ in $\mc{F}$ with $sgn(f_b(x_i)-r_i)=b_i$ for $1\leq i\leq N$.

$\mc{F}$ has pseudo-dimension $p$ if $p$ is the cardinality of the largest finite subset of $\mathbb{R}^d$ that is pseudo-shattered by $\mc{F}$. If no such largest finite subset exists, $\mc{F}$ is said to have infinite pseudo dimension.

\section{Error Bound Degradation with Bag Size}\label{app:error_bound_weakening}
The bag-to-instance generalization error bound established in Theorem \ref{thm:main1} degrades linearly with bag-size. This section provides a justification of why this degradation with bag-size is unavoidable through the example below:

Consider $D_{\mathcal{T}}$ where each instance-label in is drawn iid from $[0,1]$. Let $y_1, \dots, y_k$ be the instance-labels within a random bag B, and by construction each $y_i$ is iid and drawn u.a.r. from $[0,1]$. Using simple integration we obtain $\textnormal{E}[y_i - 1/2] = 0$ and $\textnormal{E}[(y_i - 1/2)^2] = 1/12$. Consider a regressor $h$ with a constant prediction of $1/2$. The expected loss on a random bag is $\textnormal{E}[((\sum_{i=1}^ky_i)/k - 1/2)^2] = \textnormal{E}[(\sum_{i=1}^ky_i - k/2)^2]/k^2 = 1/(12k)$. Using Chernoff bounds we obtain with high probability, that the average loss on $m$ iid sampled bags $\mathcal{B}$ satisfies $\overline{\varepsilon}(\mathcal{B}, h) \approx 1/(12k)$.  
On the other hand, the expected distributional instance-level loss is simply $\textnormal{E}[(y - 1/2)^2] = 1/12$ where $y$ is chosen u.a.r. from $[0,1]$, and thus $\varepsilon(D_{\mathcal{T}}, h) = 1/12$ and therefore one needs to incur a blowup of a factor linear in bag-size $k$.

\section{BASELINE TECHNIQUES}\label{app:baselines}
In \cite{Li-Culotta}, authors define several baselines and propose new methods for domain adaptation in LLP setting for classification tasks. We adapt these methods for regression tasks and consider those as baselines. These baselines are defined in Sections \ref{average_feature_method}, \ref{label_regularization_method}, \ref{average_feature_dann_method} and \ref{label_regularization_dann_method}. 
In literature on domain adaptation (for non-LLP settings) \citep{long2015learning, long2017deep}, it has been shown that approaches using MMD (maximum mean discrepancy) based objectives work well. Hence, we also define a baseline that uses similar objective adapted for our setting in Section \ref{domain_mean_alignment_method}.

\subsection{Average Feature Method (AF)}\label{average_feature_method}
The feature vectors in a bag are averaged and then predictions are made for the bag-averaged feature vectors via a neural network. 
The L2 loss function is used to compute difference between the predictions and bag level labels for both the source and target domain, the sum of which is used as the objective for optimization.

Let us define average bag feature by $\bar{x}_B$ such that,
\[
    \bar{x}_B = \frac{\sum\limits_{\bx \in B}\bx}{|B|}
\]
Then, the objective is defined as follows.
\[
    J(h, \mc{S}, \mc{B}) = \sum\limits_{B,y_B\in \mc{T}}\left(y_B-h\left(\bar{x}_B\right)\right)^2 + \hat{\varepsilon}(\mc{S}, h)
\]

\subsection{Label Regularization Method (LR)}\label{label_regularization_method}
This method is similar to Average Input Method with the only difference that predictions are made via neural network for each of the feature vectors in a bag first and then the predictions are averaged.
\[
    J(h, \mc{S}, \mc{B}) = \hat{\varepsilon}(\mc{S}, h) + \bar{\varepsilon}(\mc{B}, h)
\]

\subsection{Average Feature DANN Method (AF-DANN)}\label{average_feature_dann_method}
In Sections \ref{average_feature_method} and \ref{label_regularization_method}, the objective function just aimed to fit the model onto the the data from source and domain data without considering any shift in the distribution of the source and domain datasets. Average Input DANN (Domain Adversarial Neural Network) Method incorporates additional term in the Average Feature Method's objective to learn features invariant to domain and then use those features for making predictions. This is achieved by introducing an adversarial loss in form of domain prediction. The features from penultimate layer of the neural network are used to classify the input feature vector as belonging to the source/target domain. We denote this domain classifier by $h_d: x \rightarrow [0,1]$ such that $h_d(x) = \sigma(W_{h_d}^T(\phi_h(x)) + b_{h_d})$ where $\sigma$ denotes the sigmoid function and $h$ is the actual function approximator. If the classifier is not able to correctly classify labels, it means that the feature representations learnt by the network are invariant to the domain shift. The overall objective is given by $J$ as follows.
\begin{align*}
    &J(h, \mc{S}, \mc{B}) = \begin{aligned}[]\sum\limits_{B,y_B\in \mc{T}}\left(y_B-h\left(\bar{x}_B\right)\right)^2 + \hat{\varepsilon}(\mc{S}, h) -\lambda (L_D)\end{aligned}\\
    &L_D = \sum\limits_{\bx,y\in \mc{S}}\mathcal{L}(1, h_d(\bx)) + \sum\limits_{B,y_B\in \mc{T}}\sum\limits_{\bx\in B}\mathcal{L}(0, h_d(\bx))\\
    &\mathcal{L}(y,\hat{y}) = -ylog(\hat{y})-(1-y)log(1-\hat{y})
\end{align*}
We call $L_D$ the domain loss. This objective is optimized in two steps. In the first step, $J$ is minimized while keeping $(W_{h_d}$ and $b_{h_d}$ fixed. In the second step, $J$ is maximized while keeping everything but $(W_{h_d}$ and $b_{h_d}$ fixed. Essentially, in the first step encourage domain misclassifications so that the model learns feature representation that is invariant to domain shift present in the dataset. In the second step, the domain classifier is learnt for the updated feature representations. It is worth noting that the domain loss neither depends on the instance level labels from source domain nor does it depend on the bag level labels from target domain.

\subsection{Label Regularization DANN Method (LR-DANN)}\label{label_regularization_dann_method}
This method is similar to AF-DANN method (defined in Section  \ref{average_feature_dann_method}). The only difference comes from using label regularization loss instead of average feature loss in the objective function. The overall objective hence becomes as follows.
\[
    J(h, \mc{S}, \mc{T}) = \bar{\epsilon}(\mc{B}, h) + \hat{\epsilon}(\mc{S},h) - \lambda (L_D)
\]
where $L_D$ is the same as defined in Section \ref{average_feature_dann_method}.

\subsection{Domain Mean Feature Alignment Method (DMFA)}\label{domain_mean_alignment_method}
The idea is to make the feature representations domain-invariant by reducing the distance between the mean of feature representations from the source and the target domain. The overall objective is given by $J$ as follows.
\begin{align*}
    &J(h, \mc{S}, \mc{T}) = \bar{\epsilon}(h, \mc{T}) + \hat{\epsilon}(h, \mc{S}) + \lambda (L_{DMFA})\\
    &L_{DMFA} = \left\lVert\sum\limits_{B,y_B\in \mc{T}}\sum\limits_{\bx\in B}\frac{\phi(\bx)}{|B||\mc{T}|} - \sum\limits_{\bx,y\in D_S} \frac{\phi(\bx)}{|D_S|}\right\rVert_2^2
\end{align*}
Note that just like AF-DANN method (defined in Section  \ref{average_feature_dann_method}) and LR-DANN (defined in Section \ref{label_regularization_dann_method}), this method also doesn't leverage instance level source labels and bag level target labels in the objective function.

\section{DATASET PREPARATION DETAILS}\label{app:dataset}
\subsection{Synthetic Dataset}
The feature vector comprises of 64 numerical features. The label is a scalar-valued continuous variable. The feature vectors are sampled from a multi-dimensional Gaussian distribution. For the Gaussian distribution, the mean vector is itself sampled from $\mathcal{N}(0, 16)$ for source domain and $\mathcal{N}(50, 16)$ for target domain. For the experiment results presented in main paper, the co-variance matrix is a diagonal matrix where the diagonal elements are sampled from $\mathcal{N}(10, 16)$ for both the source and target domain. However, we also experiment using synthetic dataset generated with non-diagonal covariance matrix, the results for which are reported in appendix. Although the process of generating co-variance matrices is same for source and target domain, the actual covariance matrices are not the same.

As we assume co-variate shift in the source and target distribution, $p(y|x)$ is same for both distributions, hence we initialize a neural network with random weights and use that for obtaining the labels corresponding to feature vectors for both the source and target data.

The train set comprises 0.2 million instances from both source and target domain. The test set comprises 65 thousand instances from target domain.

\subsection{Wine Dataset}
Wine dataset \citep{wine_ratings, wine_reviews} is a tabular dataset with 39 boolean features indicating whether a particular word was present in the review for that wine. It also has a cardinal feature named points, which ranges between 80 (inclusive) and 100 (exclusive). The label is the price of the wine. We process feature vectors to convert all features to one hot and thus obtain a $39\times 2 + (100-80) = 98$ dimensional boolean-valued multi-hot vector as input feature vector.

The labels in the dataset are skewed. To prevent the outliers from hindering the learning process, we remove the outliers by discarding features with labels in the top 5 percentile.

We split the dataset into two different domains. The source domain comprises of wines from France and the target domain comprises of wines from all countries but France. We select France as the source domain because it has enough number of instances to qualify as a separate domain and not so many that the target domain becomes small. We run another set of experiments where Italy is chosen as the source domain and the target domain comprises of wines from all countries but Italy. The results for the former are presented in the main paper (see Table \ref{tab:wine}), and those for the later configuration are presented in the appendix (see Table \ref{tab:wine_italy}).

The train set comprises 0.5 million instances from both source and target domain. The test set comprises 0.2 million instances from target domain.

\subsection{IPUMS Dataset}
IPUMS \citep{ipums_usa_2024} is a large tabular US Census dataset with a huge number of features. For our experiments, we select income (INCWAGE) as the label. We select a subset of feature columns comprising of the following features: REGION, STATEICP, AGE, IND, GQ, SEX and WKSWORK2. All of these features are categorical except AGE which is cardinal. We convert GQ (5 categories), SEX (2 categories), WKSWORK2 (7 categories) to one-hot representations while keeping others intact as they have large number of categories which makes one-hot representations impractical.

We consider the data from 1970 as the source domain and data from 2022 as the target domain. Since, the labels (INCWAGE) were large in magnitude, we standardized the labels using $y\rightarrow (y-\mu_Y)/\sigma_Y$ by estimating the mean and variance using source domain labels and target domain train labels only.

The train set comprises 1.3 million instances from source and 9.4 million instances from target domain. The test set comprises 0.3 million instances from target domain. 

\subsection{Criteo SSCL Dataset}
Criteo Sponsored Search Conversion Log Dataset \citep{tallis2018reacting} comprises of 90 days of Criteo live traffic data. Every row in the dataset corresponds to a click (product related advertisement) that was displayed to a user. The preprocessing of the dataset is the same as done by \cite{brahmbhatt2024llp}.

We remove all the rows where the label is -1 because these instances indicate no conversion. Further, we remove all the rows where NaN or -1 is present. For our experiments, we select sales\_amount\_in\_euro as the label. The feature representation comprises of 15 categorical (product\_age\_group, device\_type, audience\_id, product\_gender, product\_brand, product\_category\_1, product\_category\_2, product\_category\_3, product\_category\_4, product\_category\_5, product\_category\_6, product\_category\_7, product\_title, partner\_id, user\_id) and 3 numerical features (time\_delay\_for\_conversion, nb\_clicks\_1week, product\_price). An embedding of dimension 8 is learnt for all the categorical features in the neural network.

The train set comprises 0.5 million instances from source and 0.9 million instances from target domain. The test set comprises 0.2 million instances from target domain. 

\section{Hyperparameter Search}\label{app:hyperparameters}
We use grid search for finding optimal values of $\lambda$ and learning rate. The values used in grid search are on a logarithmic scale. We try out two different optimizers for all experiments mentioned in the main paper - Adam and SGD and report scores corresponding to the best performer. We observed that Adam works better for most of the cases, so we perform experiments described in Appendix with Adam optimizer only.

Note that the magnitude of $\xi^2(\mc{S}, \mc{B})$ term in ${\sf BagCSI}$ loss depends on the embedding and hence the initialization of the network.
Hence, we scale $\xi^2(\mc{S}, \mc{B})$ value to match $\bar{\varepsilon}(\mc{B}, h)$.
Effectively the ${\sf BagCSI}$ contains $(\kappa \times \lambda_3)\xi^2(\mc{S}, \mc{B})$, where $\kappa = \frac{\bar{\varepsilon}(\mc{B}, h)}{\xi^2(\mc{S}, \mc{B})}$.
It must be noted that $\kappa$ is a constant and no gradient flows through it. $(\kappa \times \lambda_3)$ is an adaptive weight for ${\xi^2(\mc{S}, \mc{B})}$ term.
We do this for all methods (including baselines) that use a $\lambda$ hyperparameter.

\section{ADDITIONAL EXPERIMENTS}\label{app:results}
In addition to the experiments for which the results were shared in the main paper, we conduct a few more experiments and extensive ablation studies. The setup and results for these experiments are shared in the following sub-sections. More precisely, we perform the following experiments:
\begin{enumerate}
    \item We create a different source-target domain split in Wine dataset by choosing wines from Italy in the source domain partition and wines from all other countries in the target domain partition. The results are reported in Table \ref{tab:wine_italy}.
    \item We create another synthetic dataset where we choose a non-diagonal covariance matrix while keeping all other configurations the same. The results are reported in Table \ref{tab:synthetic_non_diagonal}.
    \item We empirically study the impact of excluding regularization term in the loss function on the performance of proposed methods. The experimental setup and results are detailed in Appendix \ref{app:excluding_regularization_term}.
    \item We also perform experiments to study the impact on performance of different algorithms by varying the amount of covariate shift in the synthetic dataset. The setup and results are detailed in Appendix \ref{app:perturbation}.
\end{enumerate}

\input{tab_reg_term_mag}

\input{tab_reg_term_mix}
\input{tab_reg_term_train}

\begin{table}[htbp]
\captionsetup{font=small,labelfont=small}
\begin{minipage}{0.465\textwidth}
\caption{MSE scores for different methods and bag sizes on the wine dataset (averaged over 20 runs) using wines from Italy as the source domain. The source instance loss is $204.73 \pm 2.7$ and target instance loss is $173.91 \pm 0.2$. Lower is better.}
\centering
\tiny
\begin{tabular}{c|c|c|c|c}\label{tab:wine_italy}
\diagbox{{Method}}{{Bag Size}} & {8} & {32} & {128} & {256} \\ \hline
Bagged Target & {176.2 $\pm$ 0.4} & {180.1 $\pm$ 0.9} & 193.8 $\pm$ 4.2 & 208.0 $\pm$ 4.5 \\
AF & 199.3 $\pm$ 2.3 & 203.2 $\pm$ 2.7 & 203.1 $\pm$ 2.5 & 203.7 $\pm$ 2.1 \\
LR & 196.0 $\pm$ 1.1 & 201.0 $\pm$ 1.0 & 203.0 $\pm$ 1.1 & 203.2 $\pm$ 0.8 \\
AF-DANN & 193.5 $\pm$ 3.5 & 195.6 $\pm$ 3.3 & 196.2 $\pm$ 3.1 & 194.7 $\pm$ 3.0 \\
LR-DANN & 195.4 $\pm$ 2.5 & 198.6 $\pm$ 3.4 & 199.7 $\pm$ 3.4 & 199.0 $\pm$ 4.1 \\
DMFA & 195.5 $\pm$ 2.2 & 201.0 $\pm$ 1.2 & 202.5 $\pm$ 1.3 & 203.2 $\pm$ 1.1 \\
PL-WFA (our) & 186.2 $\pm$ 1.0 & 188.8 $\pm$ 0.7 & 190.0 $\pm$ 0.7 & 190.3 $\pm$ 0.8 \\
BL-WFA (our) & 184.5 $\pm$ 0.6 & 187.2 $\pm$ 1.8 & {188.4 $\pm$ 1.2} & {188.0 $\pm$ 0.8} \\
\end{tabular}
\end{minipage}
\hfill
\captionsetup{font=small,labelfont=small}
\begin{minipage}{0.52\textwidth}
\caption{MSE scores for different methods and bag sizes on the synthetic dataset (averaged over 20 runs). The source instance loss is $558.3179 \pm 65.77$ and target instance loss is $9.7217 \pm 0.40$. Lower is better.}
\centering
\tiny
\vspace{3.6mm}
\begin{tabular}{c|c|c|c|c}\label{tab:synthetic_non_diagonal}
\diagbox{{Method}}{{Bag Size}} & {8} & {32} & {128} & {256} \\ \hline
Bagged Target & 29.53 $\pm$ 0.94 & 58.06 $\pm$ 1.93 & 128.45 $\pm$ 7.02 & 195.41 $\pm$ 9.34 \\
AF & 75.19 $\pm$ 3.30 & 104.36 $\pm$ 4.7 & 146.00 $\pm$ 11.7 & 207.08 $\pm$ 15.78 \\
LR & 28.36 $\pm$ 0.54 & 54.99 $\pm$ 1.74 & 120.08 $\pm$ 5.78 & 194.86 $\pm$ 11.18 \\
AF-DANN & 74.09 $\pm$ 4.13 & 107.31 $\pm$ 5.7 & 152.74 $\pm$ 24.0 & 203.54 $\pm$ 16.14 \\
LR-DANN & 30.40 $\pm$ 0.69 & 60.58 $\pm$ 2.42 & 130.30 $\pm$ 7.87 & 185.38 $\pm$ 25.97 \\
DMFA & {28.07 $\pm$ 0.63} & {54.65 $\pm$ 2.00} & 118.71 $\pm$ 7.15 & 175.68 $\pm$ 15.55 \\
PL-WFA (our) & 33.75 $\pm$ 0.67 & 63.86 $\pm$ 2.43 & 119.87 $\pm$ 5.51 & 174.12 $\pm$ 7.47 \\
BL-WFA (our) & 39.03 $\pm$ 3.73 & 65.45 $\pm$ 3.86 & {116.92 $\pm$ 17.7} & {159.08 $\pm$ 19.62} \\
\end{tabular}
\end{minipage}
\end{table}

\subsection{Experiments with Mixed Bag Sizes}\label{app:mixed_bag_sizes}
We test the performance of all the baselines and proposed methods when using a non-uniform bag size. The dataset is partitioned into bags of different sizes. More specifically, we use 2 different techniques to have mixed size bags:
\begin{itemize}
    \item \textit{SBB} is sample balanced bagging. For a particular bag size, there are an equal number of samples that belong to a bag of that size. Hence, if there are $n_1$ bags of size $k_1$, and $n_2$ bags of size $k_2$, then $n_1k_1=n_2k_2$.
    \item \textit{BBB} is bag balanced bagging. There are equal number of bags of each size. Hence, if there are $n_1$ bags of size $k_1$, and $n_2$ bags of size $k_2$, then $n_1=n_2$.
\end{itemize}
Clearly, SBB will have more bags of smaller sizes compared to BBB. Every bag is of the size 8, 32, 128 or 256. Tables \ref{tab:bbb_mixed_bag_size} and \ref{tab:sbb_mixed_bag_size} contain the results for experiments with mixed bag sizes. It can be inferred from the results that the scores with mixed bag sizes are mostly an interpolation (not necessarily linear) of the results with uniform bag sizes. Scores with BBB strategy for mixing bags are worse compared to SBB since SBB has a higher proportion of small sized bags compared to BBB.

\subsection{Experiments with Correlated Bags}\label{app:correlated_bags}
We test the performance of all the baselines and proposed methods with correlated bags as opposed to random bags used for all other experiments in this paper. To partition the dataset into correlated bags, we select a feature and create bags such that all the samples in that bag have the same value of that feature if the feature is categorical. If the feature is numerical, we sort the dataset on the basis of that feature and use consecutive samples for creating the bags.
Since all the features in Wine dataset are binary, we did not perform experiments for it. We used \textit{REGION} for IPUMS and \textit{product\_brand} for Criteo SSCL as the correlated feature. Tables \ref{tab:correlated_ipums}, \ref{tab:correlated_synthetic} and \ref{tab:correlated_criteo} contain results for experiments with correlated bags on IPUMS, Synthetic datasets and Criteo SSCL respectively. It can be inferred that the standard deviation values are very low. This is expected because across different runs, similar bags would be created unlike experiments for un-correlated bags where the bags created for each run would comprise of a different set of instances.

\subsection{Synthetic Dataset with varying Perturbations}\label{app:perturbation}
We also conduct experiments to analyze the impact on performance of different methods by varying the amount of covariate shift in the source and target domains of the synthetic datasets. The covariate shift can be controlled using the mean and standard deviation of the source and target distributions.

The $\epsilon$ parameter is a measure of the the perturbation between the mean vectors of the source and target distributions, and $\delta$ is that for the perturbation between the covariance matrices. Specifically, a target distribution is given by a 64-dimensional Gaussian where each entry is iid, sampled from $N(50, 8)$ while $\Sigma$ is a diagonal matrix where each diagonal element is the magnitude of an iid value sampled from $N(10, 8)$. For each $(\epsilon, \delta)$, the source distribution is $N(\mu', \sigma')$ where $\mu'=\mu - \epsilon\Delta$ and $\Delta$ is a vector with iid values samples from $N(50, 8)$. The diagonal matrix $\Sigma'$ is obtained by adding the magnitude of value sampled from iid $N(0, 8\delta^2)$ to each diagonal entry of $\Sigma$.

We perform experiments for different perturbations in the mean vector (using $\epsilon$) and covariance matrix (using $\delta$) of source and target distributions. As expected, with increasing perturbations, the scores become higher. Since MSE scores worsen more consistently with increase in mean perturbation as compared to perturbation in covariance matrix, we infer that the impact of increasing mean perturbation is more prominent compared to the perturbation in covariance matrix. Table \ref{tab:synth-perturbation} contains scores for different combinations of perturbation values.
 
\input{tab_synth_perturbation}

%% file: tab_reg_term_mag.tex
\begin{table*}[htbp]
\captionsetup{font=small,labelfont=small}
\begin{minipage}{0.48\textwidth}
\centering
\caption{Comparison of magnitude of the regularization term $\mathcal{R}(h, \mathcal{S}, \mathcal{T})$ and the magnitude of ${\sf BagCSI}$ loss on IPUMS dataset.}\label{tab:reg_term_mag_ipums}
\tiny
\begin{tabular}{l|rr|rr}
IPUMS &\multicolumn{2}{c}{PLWFA} &\multicolumn{2}{c}{BLWFA} \\\cmidrule{1-5}
\diagbox{{Bag Size}}{{Method}} &$\mathcal{R}(h, \mathcal{S}, \mathcal{T})$ &${\sf BagCSI}$ &$\mathcal{R}(h, \mathcal{S}, \mathcal{T})$ &${\sf BagCSI}$ \\\midrule
8 &0.01 &1.59 &0.02 &1.59 \\
32 &0.03 &1.62 &0.03 &1.62 \\
128 &0.04 &1.62 &0.05 &1.62 \\
256 &0.06 &1.62 &0.06 &1.63\\
\end{tabular}
\end{minipage}
\hfill
\begin{minipage}{0.48\textwidth}
\centering
\caption{Comparison of magnitude of the regularization term $\mathcal{R}(h, \mathcal{S}, \mathcal{T})$ and the magnitude of ${\sf BagCSI}$ loss on Criteo dataset.}\label{tab:reg_term_mag_criteo}
\tiny
\begin{tabular}{l|rr|rr}
Criteo &\multicolumn{2}{c}{PLWFA} &\multicolumn{2}{c}{BLWFA} \\\cmidrule{1-5}
\diagbox{{Bag Size}}{{Method}} &$\mathcal{R}(h, \mathcal{S}, \mathcal{T})$ &${\sf BagCSI}$ &$\mathcal{R}(h, \mathcal{S}, \mathcal{T})$ &${\sf BagCSI}$ \\\midrule
64 &17.52 &577.50 &17.44 &569.73 \\
128 &17.46 &611.63 &17.40 &602.30 \\
256 &17.97 &654.50 &17.85 &644.50 \\
512 &18.19 &673.34 &18.22 &662.61 \\
\end{tabular}
\end{minipage}
\end{table*}

%% file: tab_reg_term_mix.tex
\begin{table*}[htbp]
\captionsetup{font=small,labelfont=small}
\begin{minipage}{0.48\textwidth}
\centering
\caption{Comparison of magnitude of the regularization term $\mathcal{R}(h, \mathcal{S}, \mathcal{T})$ and the magnitude of ${\sf BagCSI}$ loss on performance for Wine dataset.}\label{tab:reg_term_mag_wine}
\tiny
\begin{tabular}{l|rr|rr}
    Wine &\multicolumn{2}{c}{PLWFA} &\multicolumn{2}{c}{BLWFA} \\\cmidrule{1-5}
    \diagbox{{Bag Size}}{{Method}} &$\mathcal{R}(h, \mathcal{S}, \mathcal{T})$ &${\sf BagCSI}$ &$\mathcal{R}(h, \mathcal{S}, \mathcal{T})$ &${\sf BagCSI}$ \\\midrule
    8 &0.66 &704.05 &0.62 &698.44 \\
    32 &1.04 &707.39 &1.14 &700.56 \\
    128 &1.32 &708.33 &1.33 &701.18 \\
    256 &1.73 &713.34 &1.75 &706.06 \\
\end{tabular}
\end{minipage}
\hfill
\begin{minipage}{0.48\textwidth}
\centering
\caption{Effect of adding the regularization term $\mathcal{R}(h, \mathcal{S}, \mathcal{T})$ to loss on performance for Wine dataset. For significant impact of the extra regularization term, the hyperparameter search is done within range $W = \{10^{-5}, 5\times 10^{-5}, 10^{-4}, 5\times 10^{-4}, 10^{-3}, 5\times 10^{-3}, 10^{-2}\}$.}\label{tab:reg_term_train_wine}
\tiny
\begin{tabular}{l|rr|rr}
    Wine &\multicolumn{2}{c}{PLWFA} &\multicolumn{2}{c}{BLWFA} \\\cmidrule{1-5}
    \diagbox{{Bag Size}}{{Method}} &$w_R = 0$ &best $w_R \in W$ &$w_R = 0$ &best $w_R \in W$ \\\midrule
    8 &183.0 $\pm$ 0.6 &255.57 $\pm$ 3.81 &180.9 $\pm$ 0.5 &255.47 $\pm$ 3.91 \\
    32 &186.6 $\pm$ 1.0 &260.85 $\pm$ 3.76 &184.6 $\pm$ 0.7 &259.80 $\pm$ 3.71 \\
    128 &189.0 $\pm$ 0.8 &270.23 $\pm$ 3.69 &186.0 $\pm$ 0.8 &270.23 $\pm$ 3.66 \\
    256 &188.9 $\pm$ 1.2 &276.27 $\pm$ 3.72 &188.9 $\pm$ 1.2 &276.25 $\pm$ 3.73 \\
\end{tabular}
\end{minipage}
\end{table*}

%% file: tab_reg_term_train.tex
\begin{table*}[ht]
\captionsetup{font=small,labelfont=small}
\begin{minipage}{0.48\textwidth}
\centering
\caption{Effect of adding the regularization term $\mathcal{R}(h, \mathcal{S}, \mathcal{T})$ to loss on performance for IPUMS dataset. For significant impact of the extra regularization term, the hyperparameter search is done within range $W = \{10^{-5}, 5\times 10^{-5}, 10^{-4}, 5\times 10^{-4}, 10^{-3}, 5\times 10^{-3}, 10^{-2}\}$.}\label{tab:reg_term_train_ipums}
\tiny
\begin{tabular}{l|rr|rr}
IPUMS &\multicolumn{2}{c}{PLWFA} &\multicolumn{2}{c}{BLWFA} \\\cmidrule{1-5}
\diagbox{{Bag Size}}{{Method}} &$w_R = 0$ &best $w_R \in W$ &$w_R = 0$ &best $w_R \in W$ \\\midrule
8 &1.15 $\pm$ 0.00 &1.23 $\pm$ 0.01 &1.14 $\pm$ 0.00 &1.23 $\pm$ 0.01 \\
32 &1.18 $\pm$ 0.00 &1.32 $\pm$ 0.02 &1.16 $\pm$ 0.00 &1.32 $\pm$ 0.02 \\
128 &1.25 $\pm$ 0.01 &1.38 $\pm$ 0.03 &1.22 $\pm$ 0.00 &1.38 $\pm$ 0.03 \\
256 &1.29 $\pm$ 0.01 &1.42 $\pm$ 0.02 &1.25 $\pm$ 0.01 &1.42 $\pm$ 0.02\\
\end{tabular}
\end{minipage}
\hfill
\begin{minipage}{0.48\textwidth}
\centering
\caption{Effect of adding the regularization term $\mathcal{R}(h, \mathcal{S}, \mathcal{T})$ to loss on performance for Criteo dataset. For significant impact of the extra regularization term, the hyperparameter search is done within range $W = \{10^{-5}, 5\times 10^{-5}, 10^{-4}, 5\times 10^{-4}, 10^{-3}, 5\times 10^{-3}, 10^{-2}\}$.}\label{tab:reg_term_train_criteo}
\tiny
\begin{tabular}{l|rr|rr}
Criteo &\multicolumn{2}{c}{PLWFA} &\multicolumn{2}{c}{BLWFA} \\\cmidrule{1-5}
\diagbox{{Bag Size}}{{Method}} &$w_R = 0$ &best $w_R \in W$ &$w_R = 0$ &best $w_R \in W$ \\\midrule
64 &204.71 $\pm$ 2.6 &256.11 $\pm$ 4.45 &204.62 $\pm$ 2.4 &255.23 $\pm$ 4.57 \\
128 &226.39 $\pm$ 2.9 &264.97 $\pm$ 3.61 &226.33 $\pm$ 2.9 &264.45 $\pm$ 3.50 \\
256 &240.55 $\pm$ 3.3 &279.45 $\pm$ 2.06 &240.39 $\pm$ 3.2 &279.48 $\pm$ 2.06 \\
512 &254.46 $\pm$ 5.5 &291.38 $\pm$ 2.44 &254.36 $\pm$ 5.5 &290.96 $\pm$ 2.46 \\
\end{tabular}
\end{minipage}
\end{table*}

%% file: tab_synth_perturbation.tex
\begin{table}[htbp]\centering
\captionsetup{font=small,labelfont=small}
\caption{Effect of covariate shift in synthetic data on the performance (MSE scores) of different algorithms for different bag sizes. Lower is better.}\label{tab:synth-perturbation}
\scriptsize
\begin{tabular}{|l|r|r|r|r|r|r|r|r|r|}\toprule
$\epsilon$ &$\delta$ &AF &LR &AFDANN &LRDANN &DMFA &PLWFA &BLWFA \\\midrule
\multicolumn{9}{c}{For bag size {8}, Bagged Target scores 0.67 $\pm$ 0.06} \\
\midrule
\multirow{2}{*}{0} &0.5 &4.56 $\pm$ 0.83 &3.93 $\pm$ 0.25 &4.85 $\pm$ 0.32 &4.58 $\pm$ 0.32 &{3.49 $\pm$ 0.45} &3.78 $\pm$ 0.24 &3.73 $\pm$ 0.39 \\
&1 &3.85 $\pm$ 0.95 &4.06 $\pm$ 1.55 &4.40 $\pm$ 0.60 &4.25 $\pm$ 0.52 &{3.36 $\pm$ 0.34} &4.14 $\pm$ 1.50 &4.06 $\pm$ 1.52 \\
\hline
\multirow{3}{*}{0.5} &0 &3.16 $\pm$ 0.27 &2.95 $\pm$ 0.42 &3.67 $\pm$ 0.21 &3.58 $\pm$ 0.21 &2.85 $\pm$ 0.49 &2.84 $\pm$ 0.45 &{2.81 $\pm$ 0.48} \\
&0.5 &2.90 $\pm$ 0.27 &2.60 $\pm$ 0.14 &3.18 $\pm$ 0.15 &3.15 $\pm$ 0.09 &2.58 $\pm$ 0.18 &{2.57 $\pm$ 0.14} &2.57 $\pm$ 0.22 \\
&1 &2.81 $\pm$ 0.43 &2.43 $\pm$ 0.20 &3.32 $\pm$ 0.57 &2.89 $\pm$ 0.21 &2.43 $\pm$ 0.18 &2.44 $\pm$ 0.23 &{2.36 $\pm$ 0.16} \\
\hline
\multirow{3}{*}{1} &0 &5.56 $\pm$ 0.74 &4.56 $\pm$ 0.24 &7.65 $\pm$ 0.63 &6.35 $\pm$ 0.75 &4.55 $\pm$ 0.26 &4.50 $\pm$ 0.30 &{4.44 $\pm$ 0.23} \\
&0.5 &5.58 $\pm$ 0.67 &4.61 $\pm$ 0.36 &7.76 $\pm$ 0.80 &6.43 $\pm$ 0.36 &4.63 $\pm$ 0.25 &{4.55 $\pm$ 0.32} &4.58 $\pm$ 0.38 \\
&1 &5.47 $\pm$ 0.79 &4.71 $\pm$ 0.31 &7.90 $\pm$ 0.79 &6.46 $\pm$ 0.57 &4.70 $\pm$ 0.35 &4.70 $\pm$ 0.35 &{4.59 $\pm$ 0.37} \\
\midrule
\multicolumn{9}{c}{For bag size {32}, Bagged Target scores 16.51 $\pm$ 2.17} \\\midrule
\multirow{2}{*}{0} &0.5 &10.07 $\pm$ 1.51 &10.07 $\pm$ 1.70 &10.18 $\pm$ 1.07 &9.69 $\pm$ 0.94 &9.57 $\pm$ 1.93 &9.22 $\pm$ 1.83 &{8.21 $\pm$ 1.04} \\
&1 &10.47 $\pm$ 2.21 &10.23 $\pm$ 2.09 &9.91 $\pm$ 1.21 &9.65 $\pm$ 1.22 &9.85 $\pm$ 1.82 &9.29 $\pm$ 2.02 &{9.19 $\pm$ 2.09} \\
\hline
\multirow{3}{*}{0.5} &0 &8.31 $\pm$ 2.03 &7.58 $\pm$ 0.95 &8.53 $\pm$ 1.63 &7.28 $\pm$ 0.79 &7.49 $\pm$ 1.01 &6.61 $\pm$ 0.96 &{6.54 $\pm$ 0.98} \\
&0.5 &7.60 $\pm$ 1.66 &7.12 $\pm$ 0.63 &7.74 $\pm$ 2.06 &6.88 $\pm$ 0.78 &7.07 $\pm$ 0.52 &{6.05 $\pm$ 0.55} &6.07 $\pm$ 0.47 \\
&1 &8.27 $\pm$ 1.82 &6.62 $\pm$ 0.60 &7.87 $\pm$ 1.75 &6.81 $\pm$ 0.78 &6.68 $\pm$ 0.58 &{5.63 $\pm$ 0.60} &5.71 $\pm$ 0.73 \\
\hline
\multirow{3}{*}{1} &0 &12.43 $\pm$ 1.03 &10.18 $\pm$ 0.70 &14.06 $\pm$ 0.80 &11.11 $\pm$ 0.74 &10.04 $\pm$ 0.55 &{8.97 $\pm$ 0.52} &9.16 $\pm$ 0.76 \\
&0.5 &12.96 $\pm$ 1.18 &10.50 $\pm$ 0.80 &14.46 $\pm$ 0.56 &11.59 $\pm$ 0.70 &10.33 $\pm$ 0.89 &{9.22 $\pm$ 0.92} &9.33 $\pm$ 0.95 \\
&1 &12.85 $\pm$ 1.52 &10.49 $\pm$ 1.30 &14.45 $\pm$ 1.16 &12.15 $\pm$ 1.00 &10.23 $\pm$ 1.23 &9.27 $\pm$ 1.10 &{9.15 $\pm$ 0.98} \\
\midrule
\multicolumn{9}{c}{For bag size {128}, Bagged Target scores 18.52 $\pm$ 2.01} \\\midrule
\multirow{2}{*}{0} &0.5 &15.86 $\pm$ 1.35 &16.21 $\pm$ 2.08 &15.54 $\pm$ 1.42 &14.74 $\pm$ 1.45 &15.11 $\pm$ 1.17 &{15.03 $\pm$ 1.91} &15.11 $\pm$ 1.96 \\
&1 &16.28 $\pm$ 1.45 &15.85 $\pm$ 2.16 &15.43 $\pm$ 1.22 &15.13 $\pm$ 1.76 &15.34 $\pm$ 1.43 &14.92 $\pm$ 2.19 &{14.62 $\pm$ 1.95} \\
\hline
\multirow{3}{*}{0.5} &0 &13.82 $\pm$ 1.71 &12.39 $\pm$ 0.93 &11.91 $\pm$ 2.78 &12.60 $\pm$ 0.57 &12.34 $\pm$ 0.90 &{11.32 $\pm$ 0.96} &11.41 $\pm$ 0.89 \\
&0.5 &13.43 $\pm$ 3.21 &12.08 $\pm$ 0.83 &12.74 $\pm$ 1.42 &12.01 $\pm$ 0.86 &12.06 $\pm$ 0.82 &11.08 $\pm$ 0.83 &{10.99 $\pm$ 0.84} \\
&1 &14.54 $\pm$ 1.87 &11.64 $\pm$ 0.96 &12.39 $\pm$ 1.69 &11.65 $\pm$ 0.84 &11.65 $\pm$ 0.98 &{10.56 $\pm$ 0.95} &10.60 $\pm$ 0.97 \\
\hline
\multirow{3}{*}{1} &0 &17.98 $\pm$ 1.90 &15.15 $\pm$ 1.84 &17.09 $\pm$ 1.27 &15.27 $\pm$ 0.43 &14.76 $\pm$ 1.43 &{13.74 $\pm$ 1.50} &13.86 $\pm$ 1.42 \\
&0.5 &17.57 $\pm$ 2.34 &15.33 $\pm$ 1.30 &17.45 $\pm$ 1.15 &15.85 $\pm$ 1.59 &15.38 $\pm$ 1.78 &{13.91 $\pm$ 1.12} &14.45 $\pm$ 2.08 \\
&1 &17.64 $\pm$ 2.21 &16.30 $\pm$ 2.73 &17.23 $\pm$ 1.05 &15.75 $\pm$ 1.17 &15.26 $\pm$ 1.74 &14.42 $\pm$ 1.63 &{14.38 $\pm$ 1.66} \\
\midrule
\multicolumn{9}{c}{For bag size {256}, Bagged Target scores 19.28 $\pm$ 1.91} \\\midrule
\multirow{2}{*}{0} &0.5 &17.84 $\pm$ 1.72 &17.89 $\pm$ 1.82 &16.97 $\pm$ 0.95 &16.97 $\pm$ 1.25 &17.41 $\pm$ 1.20 &{16.78 $\pm$ 1.71} &16.86 $\pm$ 1.71 \\
&1 &17.95 $\pm$ 1.59 &17.99 $\pm$ 2.18 &17.09 $\pm$ 1.16 &16.44 $\pm$ 1.40 &18.09 $\pm$ 1.83 &16.99 $\pm$ 2.16 &{16.98 $\pm$ 2.17} \\
\hline
\multirow{3}{*}{0.5} &0 &16.43 $\pm$ 1.71 &15.06 $\pm$ 0.91 &14.97 $\pm$ 1.14 &14.76 $\pm$ 0.79 &15.08 $\pm$ 0.83 &14.04 $\pm$ 0.86 &{13.99 $\pm$ 0.96} \\
&0.5 &15.93 $\pm$ 2.33 &15.02 $\pm$ 0.52 &15.18 $\pm$ 0.88 &14.47 $\pm$ 1.24 &14.97 $\pm$ 0.57 &14.03 $\pm$ 0.56 &{13.96 $\pm$ 0.51} \\
&1 &16.44 $\pm$ 1.87 &14.27 $\pm$ 0.94 &14.58 $\pm$ 1.47 &14.03 $\pm$ 0.79 &14.23 $\pm$ 0.93 &13.29 $\pm$ 0.91 &{13.18 $\pm$ 0.94} \\
\hline
\multirow{3}{*}{1} &0 &18.46 $\pm$ 2.10 &16.58 $\pm$ 1.99 &17.47 $\pm$ 1.43 &17.22 $\pm$ 1.37 &16.83 $\pm$ 1.99 &15.68 $\pm$ 1.90 &{15.32 $\pm$ 1.48} \\
&0.5 &18.24 $\pm$ 2.15 &17.57 $\pm$ 2.34 &17.58 $\pm$ 1.13 &17.62 $\pm$ 1.31 &16.86 $\pm$ 1.40 &{16.02 $\pm$ 1.47} &16.06 $\pm$ 1.65 \\
&1 &18.53 $\pm$ 2.16 &17.31 $\pm$ 2.50 &17.95 $\pm$ 1.21 &17.31 $\pm$ 1.20 &17.18 $\pm$ 2.31 &{15.91 $\pm$ 1.90} &16.20 $\pm$ 2.35 \\
\bottomrule
\end{tabular}
\end{table}